\documentclass[preprint]{elsarticle}
\usepackage{hyperref}
\usepackage{adjustbox}
\usepackage{dingbat}
\begin{document}

\begin{frontmatter}

\title{An Unsupervised Domain Adaptation Scheme for Single-Stage Artwork Recognition in Cultural Sites}

\author[1]{Giovanni Pasqualino\fnref{fn1}} 
\author[1]{Antonino Furnari\fnref{fn1}}
\author[2]{\\Giovanni Signorello}
\author[1,2,3]{Giovanni Maria Farinella\fnref{fn1}\corref{cor1}}
\fntext[fn1]{These authors are co-first authors and contributed equally to this work.}
\cortext[cor1]{Corresponding author: Tel: +39 095 738 3205; Fax: +39 095 330 094;}
\ead{gfarinella@dmi.unict.it}

\address[1]{Department of Mathematics and Computer Science, University of Catania, Italy}
\address[2]{CUTGANA, University of Catania, Italy}
\address[3]{ICAR-CNR, National Research Council, Palermo,Italy}

\begin{abstract}
Recognizing artworks in a cultural site using images acquired from the user's point of view (First Person Vision) allows to build interesting applications for both the visitors and the site managers. However, current object detection algorithms working in fully supervised settings need to be trained with large quantities of labeled data, whose collection requires a lot of times and high costs in order to achieve good performance. Using synthetic data generated from the 3D model of the cultural site to train the algorithms can reduce these costs. On the other hand, when these models are tested with real images, a significant drop in performance is observed due to the differences between real and synthetic images. In this study we consider the problem of Unsupervised Domain Adaptation for object detection in cultural sites. To address this problem, we created a new dataset containing both synthetic and real images of 16 different artworks. We hence investigated different domain adaptation techniques based on one-stage and two-stage object detector, image-to-image translation and feature alignment. Based on the observation that single-stage detectors are more robust to the domain shift in the considered settings, we proposed a new method which builds on RetinaNet and feature alignment that we called DA-RetinaNet. The proposed approach achieves better results than compared methods on the proposed dataset. It also obtains better results on Cityscapes.
To support research in this field we release the dataset at the following link \url{https://iplab.dmi.unict.it/EGO-CH-OBJ-UDA/} and the code of the proposed architecture at \url{https://github.com/fpv-iplab/DA-RetinaNet}.
\end{abstract}

\end{frontmatter}

\section{Introduction}
\begin{figure}[t]
            \centering
            \includegraphics[width=0.49\textwidth]{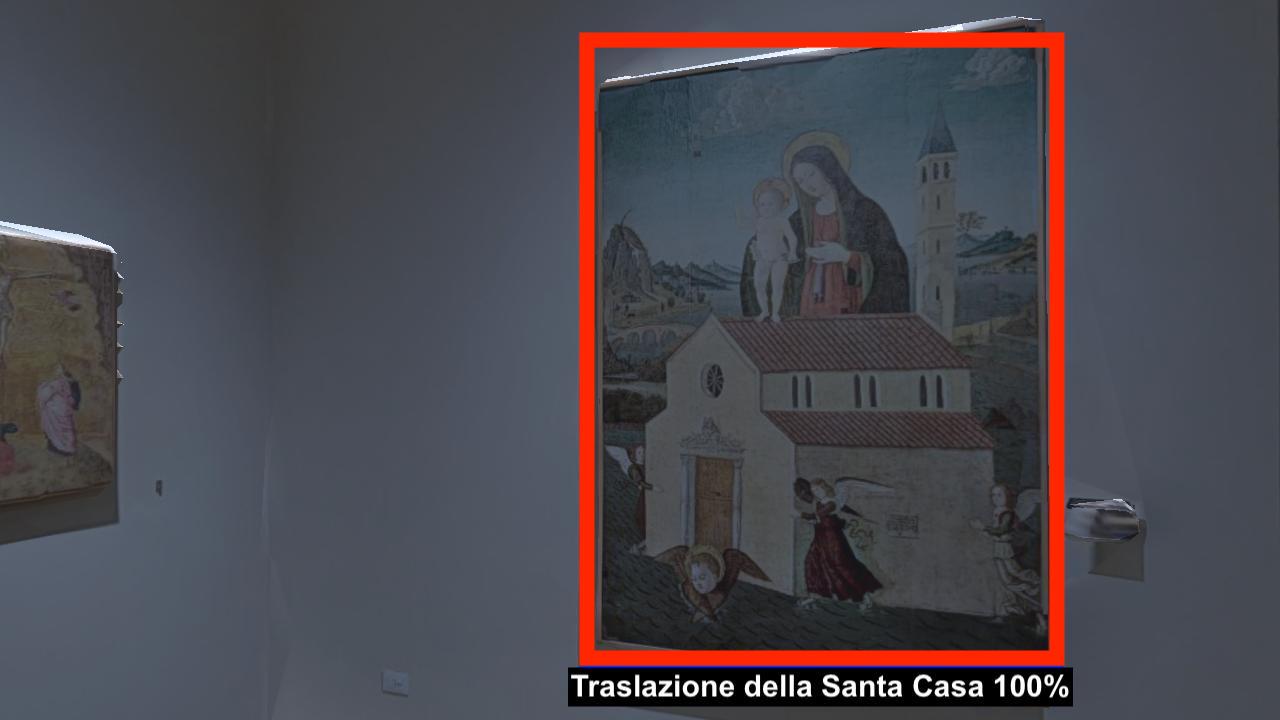}
            \includegraphics[width=0.49\textwidth]{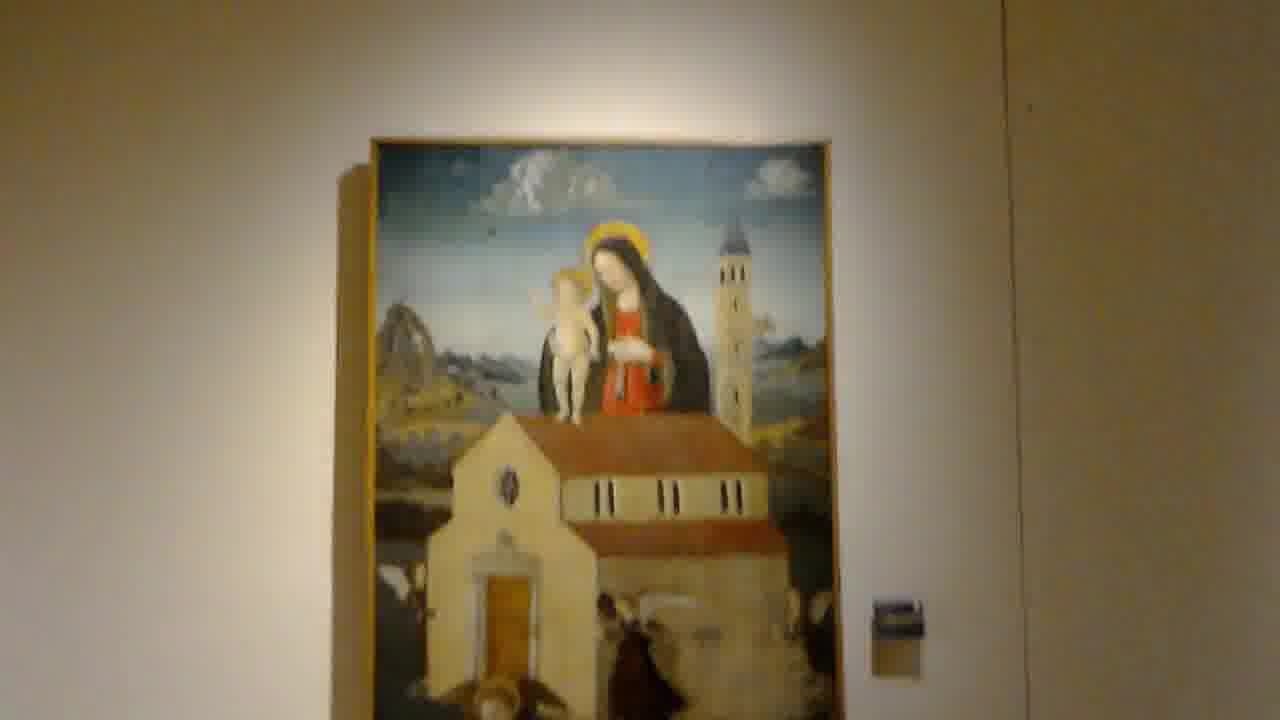}
            \caption{Example of a synthetic image generated using the tool proposed in~\cite{orlando2020egocentric} (left) and a real image of the same artwork (right).}
            \label{fig:noadapt1}
\end{figure}

Recognizing artworks in a cultural site is a key feature for many applications aimed either to provide additional services to the users or to obtain insights into the behavior of the visitors, and hence measure the performance of the cultural site~\cite{Ragusa_2020}. For example, artwork recognition allows to automatically show additional information about an artwork observed by the visitor through augmented reality~\cite{10.1145/3092832}, or monitor visitor behavior to understand where people spend more time during their visit, as well as to infer which artworks attract their interest~\cite{ragusa2019egocentricpoint}. Artwork recognition can be obtained fine-tuning standard object detector architectures (e.g. Faster-RCNN~\cite{DBLP:journals/corr/RenHG015}, YOLO~\cite{7780460}, RetinaNet~\cite{DBLP:journals/corr/abs-1708-02002}) on labeled data. However, in order to achieve good performance, object detection algorithms need to be trained on large datasets of manually labeled images. Depending on the cultural site, collecting and labeling visual data can be difficult especially when many artworks are present whose images should be acquired from different points of view. Moreover, labeling these data with bounding box annotations for each artwork is expensive and, since objects must be recognized at the instance level, the collection and labeling efforts must be repeated for each cultural site. \newline
To mitigate the aforementioned problems, a recent work~\cite{orlando2020egocentric} proposed an approach to generate large quantities of synthetic images from the 3D model of a cultural site simulating a visitor navigating the site. Since the position of artworks can been labeled in the 3D model (i.e, one 3D bounding box per artwork), all images during the simulated navigation can be automatically labeled with 2D bounding box annotations. This approach allows to easily generate labeled datasets of arbitrary size which can be used to train an object detection algorithm. Nonetheless, there is a domain gap between the generated and real visual data which the object detector models must deal with at test time. Figure~\ref{fig:noadapt1} shows  the  results  of  a  standard  Faster-RCNN  model  trained  on  the labeled synthetic images.  Due to the domain gap,  the model successfully detects artworks on synthetic images, whereas it fails on real images. Infact, these algorithms assume that the images used for training and those on which the algorithm will be tested belong to the same domain distribution. In this context, for example, if an object detector is trained with a dataset of synthetic images and tested on a dataset containing their real counterparts, the performance will drastically drop and, in many cases, the algorithm will not be able to recognize the artworks, as it shown in Figure~\ref{fig:noadapt1}.
In this experimental scenario, the set of training data is generally referred to as ``source domain", whereas the set of test data is called ``target domain". The drop in performance due to domain gap represents a significant limitation since it requires the creation of a dataset of annotated images belonging to the target domain in order to re-train or fine-tune the algorithms. The labeling process, in particular, imposes additional costs in terms of money and time. For this reason, many works have focused on reducing the domain gap by leveraging labeled images belonging to a source domain and only unlabeled images from the target domain. This research area is referred to as ``Unsupervised Domain Adaptation"~\cite{Tzeng_2017},~\cite{ganin2014unsupervised}.
\newline
In this paper, we investigate the use of unsupervised domain adaptation techniques for artwork detection. Specifically, we consider a scenario in which large quantities of labeled synthetic images are available, whereas only unlabeled real images can be used at training time. The synthetic images can be easily obtained starting from a 3D model of the cultural site acquired with a 3D scanner such as Matterport~\footnote{\url{https://matterport.com/}} and using the tool proposed in~\cite{orlando2020egocentric} to automatically generate the labeled data. The real unlabeled images can be easly collected visiting the cultural site acquiring videos with a wearable camera. Note that, since no manual labeling is required for the real images in the unsupervised settings, this procedure has a low cost. We hence aim to train the object detection models using labeled synthetic images and real unlabeled images. To the best of our knowledge, there are not publicly available datasets to study domain adaptation for artwork detection in cultural sites. Therefore we collect and publicly release a suitable one which we name UDA-CH (Unsupervised Domain Adaptation on Cultural Heritage). We hence study the main unsupervised domain adaptation techniques for object detection on UDA-CH: 1) image-to-image translation and 2) feature alignment. We compare the performance of two popular object detection approaches, Faster R-CNN~\cite{DBLP:journals/corr/RenHG015} and RetinaNet~\cite{DBLP:journals/corr/abs-1708-02002}. Since in our study RetinaNet obtained results more robust to the domain gap than Faster-RCNN, we propose a novel approach which combines feature alignment techniques based on adversarial learning~\cite{ganin2014unsupervised} for unsupervised domain adaptation with the RetinaNet architecture. Our experiments show that the proposed approach greatly outperforms prior art. When combined with image to image translation, our method achieves a mAP of $58.01\%$ on real data without seeing a single labeled real images at training time. To better demonstrate the effectiveness of the proposed method, we have also tested the generalization of the approach in urban scenario exploiting the popular Cityscapes dataset~\cite{cordts2016cityscapes},~\cite{sakaridis2018semantic}.
\newline
In sum, the contributions of this paper are as follows: 1) we introduce a new dataset to study synthetic to real unsupervised domain adaptation for artwork detection in cultural sites. The dataset has been acquired from a first person point of view on a real cultural site with 16 artworks; 2) we benchmark different solutions to address unsupervised domain adaptation for artwork detection; 3) we propose a novel architecture based on RetinaNet which obtains better results than similar approaches based on Faster-RCNN. The code of our approach is publicly available at the following link \url{https://github.com/fpv-iplab/DA-RetinaNet}; 4) We demonstrate the generalization of the proposed approach considering also a popular dataset of a different domain (i.e urban domain); 5) we analyze the limits of the investigated techniques and discuss future research directions. \newline
The remainder of the paper is organized as follows. In Section~\ref{related}, we discuss related work. Section~\ref{methods} presents the compared methods. Section~\ref{result} reports the experimental settings and discusses results. Section~\ref{conclusion} concludes the paper and summarises the main findings of our study.
\section{Related Works}
\label{related}
Our work is related to different lines of research: egocentric vision in cultural sites, object detection, image to image translation, feature alignment for domain adaptation, and unsupervised domain adaptation for object detection. The following sections discuss the relevant works belonging to these research lines.
\subsection{Egocentric vision in cultural sites}
Wearable devices can be used to improve the fruition of artworks and the user experience in cultural sites~\cite{articlecucchiara}. The authors of~\cite{10.1145/3092832},~\cite{portaz} propose a smart audio guide that, based on the actions and interests of museum visitors, interacts with the visitors improving their experience and the fruition of multimedia materials. An important ability for these systems is related to the detection of artworks, which can be achieved using object detectors~\cite{DBLP:journals/corr/RenHG015},~\cite{DBLP:journals/corr/abs-1708-02002}. Unfortunately, object detectors need to be trained with big datasets of labeled images which can be expensive to collect in cultural sites. Consequently, few datasets are available to study the problem in this context. One of the few has been proposed by the authors of~\cite{Ragusa_2020}, who collected and labeled a dataset of first person images with Microsoft Hololens in two cultural sites located in Italy. However, since artwork recognition needs to be performed at the instance level, the data collection process and the training of the algorithms has to be repeated for every cultural site. To reduce data collection and annotation costs, the authors of~\cite{orlando2020egocentric} proposed a tool to generate synthetic labeled images from a 3D reconstruction of a real cultural site. However, the generated data are not as photorealistic as the real images on which the object detection algorithm has to work at inference time, which induces a significant domain gap. Our work focuses on filling this domain gap by designing algorithms to detect objects in a cultural site considering only labeled synthetic images and unlabeled real images for training.
\subsection{Object Detection}
The detection of objects is one of the most important challenges in computer vision with impact on many aplications~\cite{6745491}, \cite{6248010}, \cite{Chen_2017}. Modern object detection algorithms are based on deep learning and can be divided into two main categories according to their architecture: the algorithms belonging to the ``two-stage" category, whose main representative are Faster R-CNN~\cite{DBLP:journals/corr/RenHG015}, Cascade R-CNN~\cite{Cai_2018} and Mask-RCNN~\cite{He_2017} and those belonging to the ``single-stage" category such as RetinaNet~\cite{DBLP:journals/corr/abs-1708-02002}, SSD~\cite{Liu_2016} and YOLO~\cite{7780460}. The former address object detection by first extracting a set of object proposal and then processing them to determine the object class and refine its position in the image. These algorithms are generally characterized by a higher accuracy in the recognition and classification of objects, but also involve higher computational costs. The latter perform object detection in a single forward pass and are characterized by a higher computational efficiency generally obtained at the expense of detection precision. In our work we compare the main representatives of the two categories, Faster-RCNN (two-stage) and RetinaNet (single-stage). We find that RetinaNet is less sensitive to the domain gap and obtains better performances.
\subsection{Domain Adaptation}
Domain adaptation is a branch of machine learning that studies solutions to adapt a model trained on a set of images following a certain distribution (source distribution), to work on a set of test images following a different distribution (target distribution). When the adaptation is done using labeled images from the source domain and unlabeled images from the target domain, the task is referred as ``Unsupervised Domain Adaptation". The authors of~\cite{Rozantsev_2019},~\cite{Sun_2016},~\cite{Kang_2019} proposed to minimize divergence quantities that can be measured between source and target distributions. Minimizing these quantities allows the model to extract features that are invariant with respect to the two domain distributions. In particular, the authors of~\cite{Rozantsev_2019} exploited the use of the MMD metric~\cite{gretton2008kernel} in a CNN to reduce the distribution mismatch. The authors of~\cite{Sun_2016} used the CORAL metrics~\cite{Sun_2017} inside a CNN to align the covariances of the source and target distributions. The authors of~\cite{Kang_2019} proposed a method that aims to minimize the intra-class discrepancy and maximize the inter-class discrepancy. Other works used adversarial learning to align the distributions of the features extracted by the models of the source and target domains. The authors of~\cite{ganin2014unsupervised} introduce a gradient reversal layer into a standard CNN to align the distributions of source and target features using adversarial learning. Specifically, the model they propose includes two components. The first one processes the input samples to solve the supervised task (e.g., classification). The second one is devoted to discriminate if the features extracted from the input sample belong to the source or target domain. The network is trained to minimize the supervised loss of the first component and the discriminator loss of the second one. The gradient reversal layer is used to invert the gradients of the discriminator when they are used to update the parameters of the first component, which implements a minmax game similar to the one described in~\cite{goodfellow2014generative}. The authors of~\cite{Tzeng_2017} propose a method based on two stages: in the first stage a CNN is trained on the source dataset. In the second stage the weights of the CNN are adapted to extract domain-invariant features. During the test phase, the weights obtained during the second stage are used to extract the features, whereas the classification layers are obtained from the network trained on the source domain.
\newline
In our work, we consider adversarial learning for domain adaptation with gradient reversal layer~\cite{ganin2014unsupervised}.
\subsection{Image-to-image translation}
When the images of the source and target domains are visually different (e.g., images acquired with different light conditions), a way to reduce the domain gap between the two domains is to use image-to-image translation techniques~\cite{Anoosheh_2019},~\cite{Murez_2018},~\cite{gonzalezgarcia2018imagetoimage}. The goal of these techniques is to translate an image belonging to a source domain into an image belonging to the target domain without changing its content but adapting only its style and colors. When pairs of images belonging to the source and target domains are available, a mapping between the two domains can be learned exploiting a conditional adversarial network~\cite{Isola_2017}. The authors of~\cite{CycleGAN2017} note that paired datasets are difficult to obtain in practice and introduce a method that translates images from a source domain $X$ to a target domain $Y$ in the absence of paired examples. As proposed in~\cite{CycleGAN2017}, the goal is to learn a function $G:X \longrightarrow Y$ such that the distribution of the transformed images $G \left (X \right)$  is indistinguishable from the distribution of $Y$. Since the translation between the two domains should be consistent, an inverse mapping $F:Y \longrightarrow X$ is introduced such that $F(G(X))\approx\ X$.
\newline
As discussed in previous works~\cite{hoffman2017cycada},~\cite{Saito_2019},~\cite{kim2019diversify}, the algorithms described in this section can be used in combination with the domain adaptation techniques of the previous subsections to deal with the domain gap. The images belonging to the source domain can be translated into the target domain and subsequently used as training images. The resulting model can be used directly on the target domain at test time. Vice versa, it is possible to train the model on the source domain and translate the test images to the source domain at inference time.
\newline
In our work, we explore the benefits of image to image translation techniques and their combination with the feature alignment methods.
\subsection{Unsupervised Domain Adaptation for object detection}
Previous works have investigated the application of unsupervised domain adaptation to the problem of object detection. Many of these works use adversarial learning with the gradient reversal layer. The authors of~\cite{chen2018domain} present a custom version of a Faster RCNN~\cite{DBLP:journals/corr/RenHG015} that includes two modules: the first one aligns the features of the entire input (i.e., at the image level), the second module aligns the features before they are used for classification and regression (i.e. at the instance level). The authors of~\cite{Saito_2019} propose to adapt source and target domains exploiting both high-and low-level features. The authors of~\cite{Xie_2019} propose an architecture similar to the one presented in~\cite{chen2018domain}, but they add more discriminators with a gradient reversal layer to the Faster-RCNN backbone. The authors of~\cite{Zhu_2019_CVPR} propose a framework to align the source and target domains at the level of image regions extracted from the ``region proposal network" of a Faster-RCNN. This architecture has two main components: 1) region mining, which extracts the regions of interest from the source and target images, groups them and selects the most important regions containing the objects; 2) the region level alignment, which learns to align the patches of the reconstructed images starting from the features selected by the previous module through adversarial learning. \newline
Recent methods address domain adaptation employing Faster-RCNN as baseline object detection architecture, whereas few approaches have investigated the use of single-stage object detectors such as RetinaNet~\cite{DBLP:journals/corr/abs-1708-02002}. In our work, we compare the performance of the two methods and introduce an object detector based on RetinaNet which includes a domain adaptation component.
\section{Methods}
\label{methods}
We compare several approaches to unsupervised domain adaptation for object detection. Specifically we considered the following: 1) a baseline object detector without adaptation, 2) domain adaptation through image-to-image translation, 3) domain adaptation through feature alignment, 4) the proposed method based on RetinaNet and feature alignment and 5) approaches combining feature alignment and image-to-image translation. In the following section, we give details on all the compared approaches.
\subsection{Baseline approaches without adaptation}
\begin{figure}[t]
            \centering
            \includegraphics[width=1\textwidth]{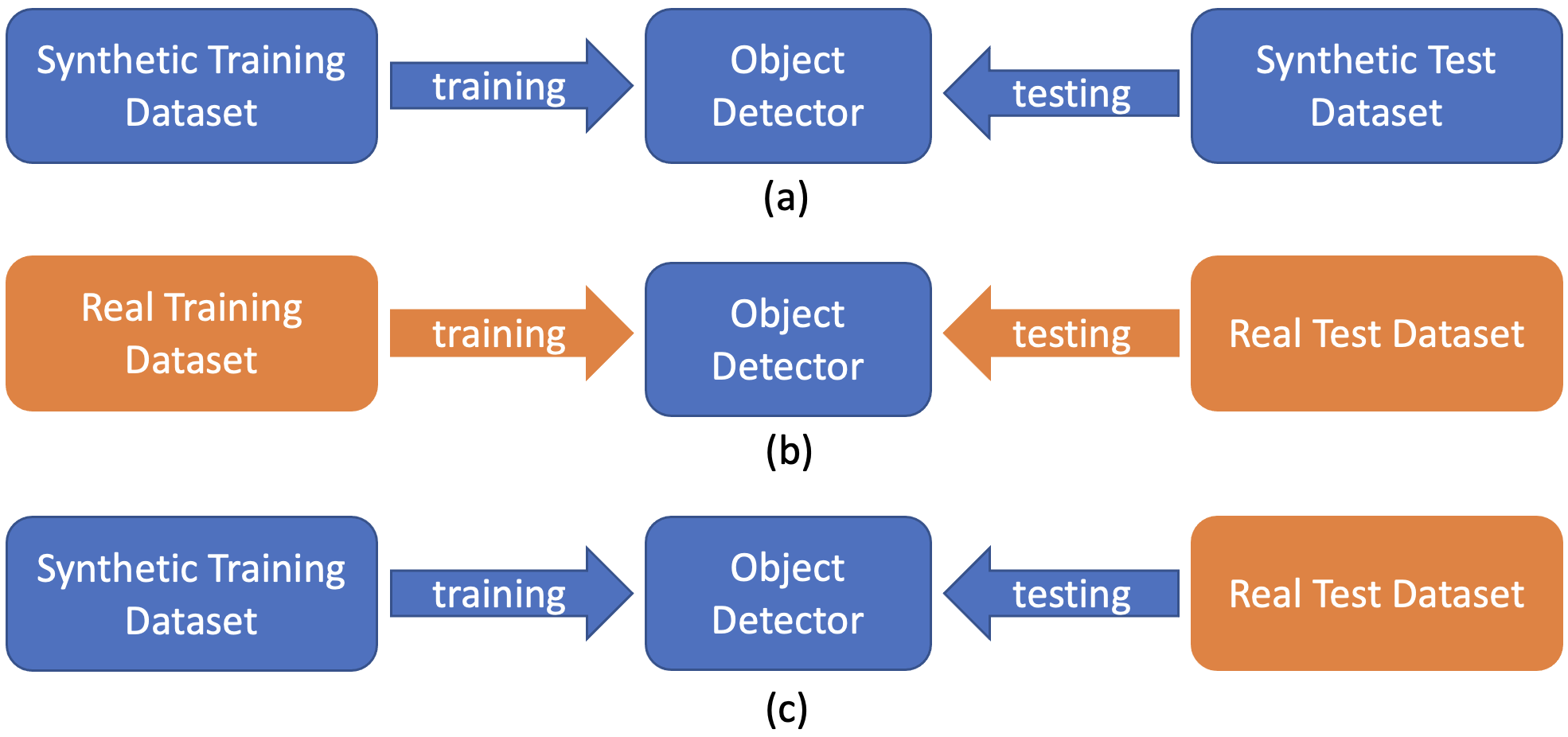}
            \caption{We used 3 different pipelines: (a) training and testing on the synthetic domain, (b) training and testing on the real domain, (c) training using synthetic images and testing on real images.}
            \label{fig:baseline}
\end{figure}

To assess performance in the absence of domain shift, we train and test Faster RCNN and RetinaNet on the same domain (either synthetic or real images), as illustrated in Figure~\ref{fig:baseline}(a) and Figure~\ref{fig:baseline}(b). We also consider a model trained on synthetic images and tested directly on real test images, as illustrated in Figure~\ref{fig:baseline}(c). These methods allow to assess the gap between the two domains.
\subsection{Domain adaptation through image-to-image translation}
\begin{figure}[t]
            \centering
            \includegraphics[width=1\textwidth]{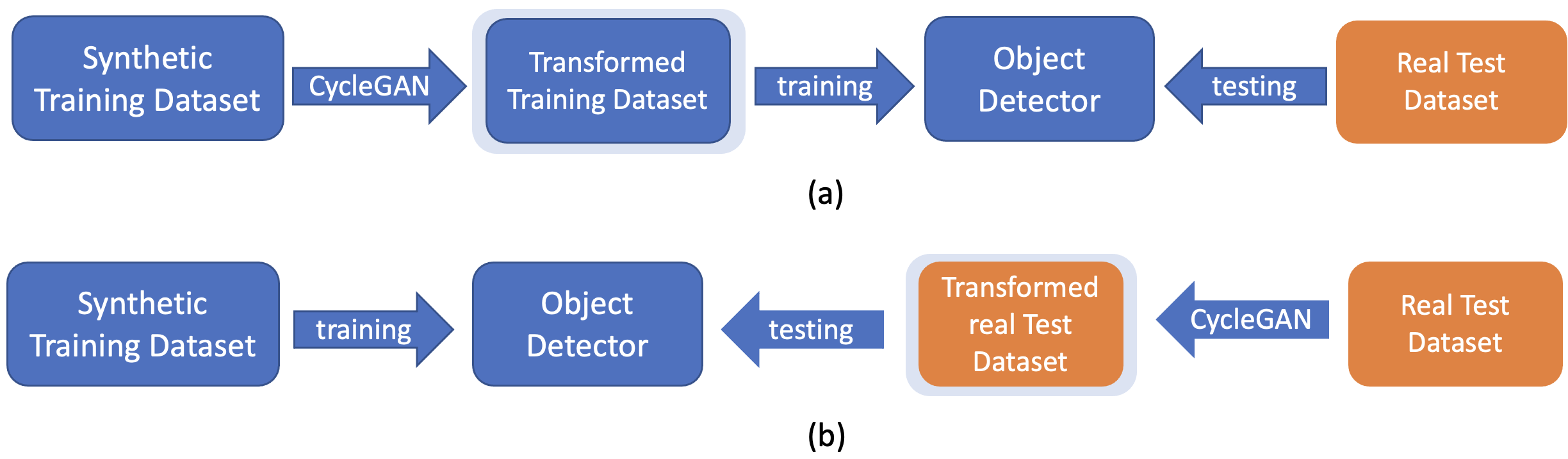}
            \caption{(a) Pipeline used to train models on synthetic images transformed to
real with test performed on real images. (b) Pipeline used to train models on synthetic images with test performed on real images transformed to synthetic.}
\label{fig:image_to_image_pip}
\end{figure}

Transforming images from synthetic to real and vice versa is a common way to reduce the domain gap. In particular, we use CycleGAN~\cite{CycleGAN2017} to transform images from one domain to another. We compare two approaches: 1) translating synthetic images to real, training Faster RCNN and RetinaNet on the transformed images and testing the two detectors with real images. This approach is illustrated in~Figure~\ref{fig:image_to_image_pip}(a); 2) translating real test images to synthetic, testing the two models that were previously trained on synthetic images as illustrated in~Figure~\ref{fig:image_to_image_pip}(b).

\subsection{Domain adaptation through feature alignment}
\label{featureAlignment}
\begin{figure}[t]
            \centering
            \includegraphics[width=1\textwidth]{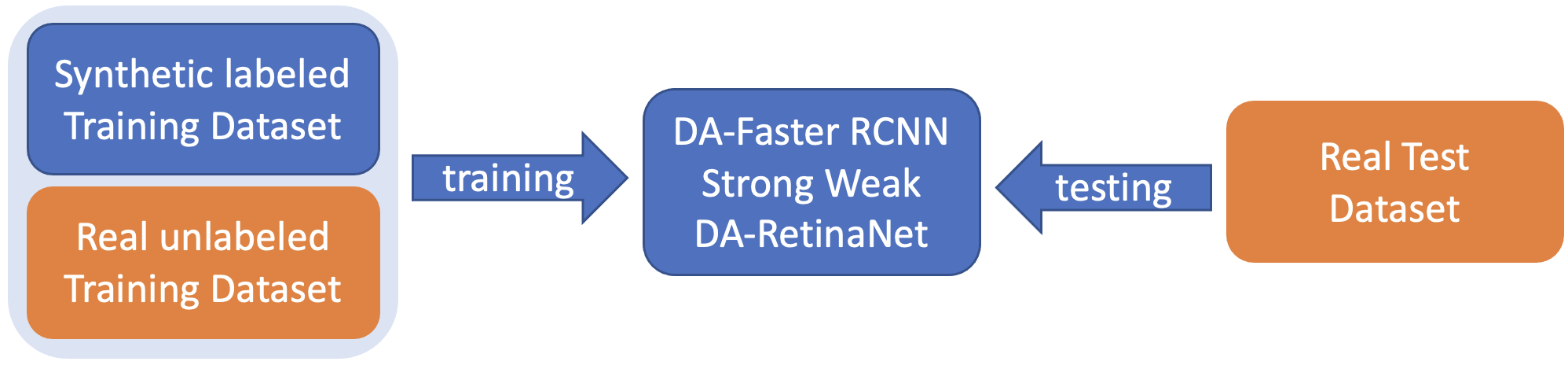}
            \caption{Pipeline used to train models based on feature alignment.}
            \label{fig:fa}
\end{figure}

We consider DA-Faster-RCNN~\cite{chen2018domain} and Strong-Weak~\cite{Saito_2019} and compare their results with our method DA-RetinaNet described in the next subsection. All these methods use synthetic labeled images and unlabeled real images for training as shown in Figure~\ref{fig:fa}.
\subsection{Proposed Method: DA-RetinaNet}
\label{DAretinanet}
\begin{figure}[t]
    \centering
	\includegraphics[width=1\linewidth]{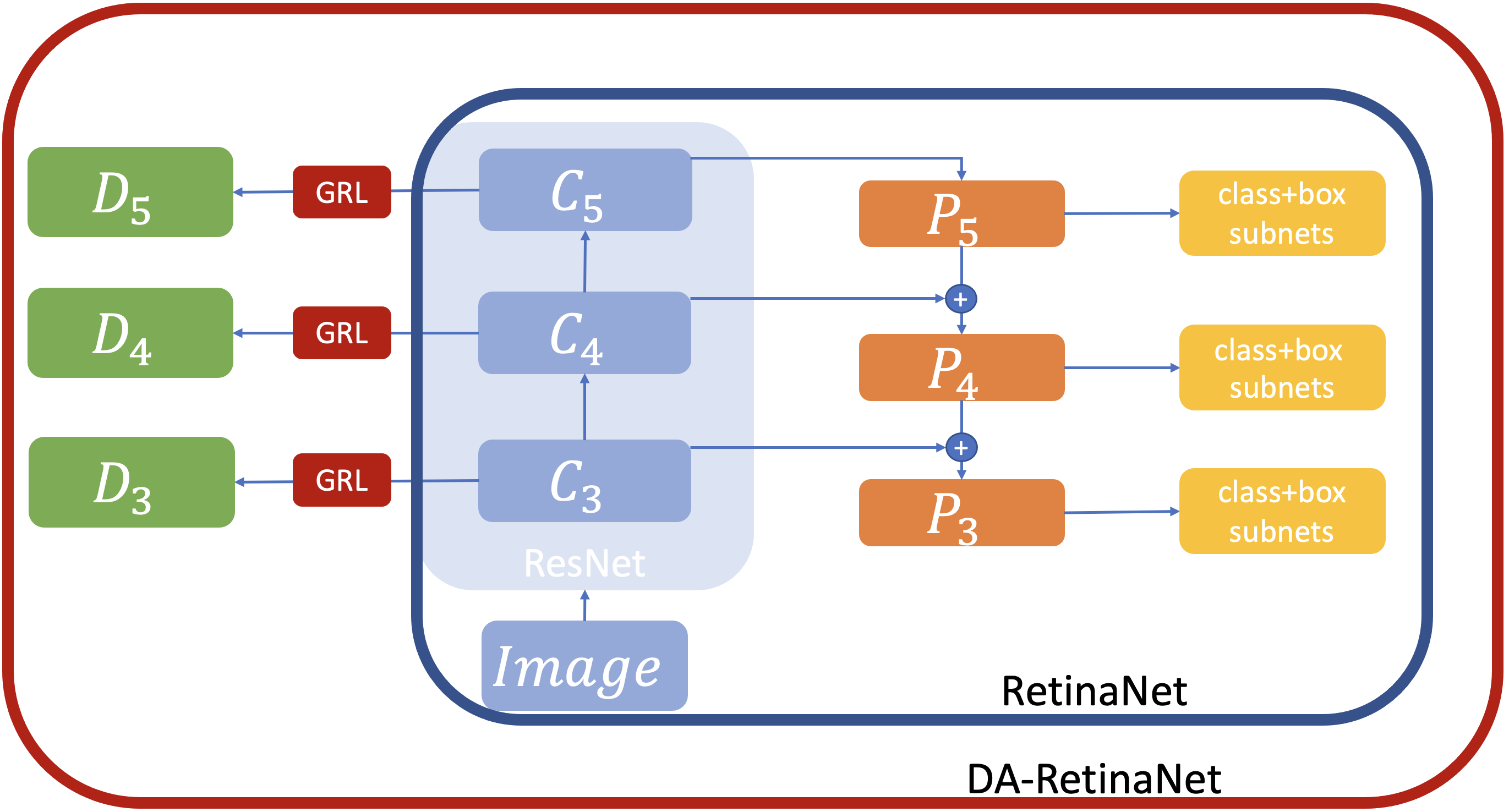} 
	\caption{Architecture of the proposed DA-RetinaNet.}
	\label{fig:DA-Retinanet} 
\end{figure}

The proposed method is based on RetinaNet architecture~\cite{DBLP:journals/corr/abs-1708-02002} and it is illustrated in Figure~\ref{fig:DA-Retinanet}. At each level of the feature pyramid map ($C_3$, $C_4$ and $C_5$) in the ResNet backbone, we add a discriminator ($D_3$, $D_4$, $D_5$) with a Gradient Reversal Layer. The three discriminators have different architectures: $D_3$ has 3 convolutional layers with a kernel size of 1 and ReLU as activation function; $D_4$ has 3 convolutional layers with kernel size of 3 followed by batch normalization, ReLU and Dropout. At the end of the last convolutional layer there is a fully connected layer; $D_5$ has 3 convolutional layers with kernel size of 3 followed by batch normalization, ReLU and Dropout. After the convolutional layer there are 2 fully connected layers. Our idea follows~\cite{ganin2014unsupervised}, thus we train our model to minimize the cost function:
\begin{equation}
    L=L_{class}+L_{box}-\lambda(L_{D3}+L_{D4}+L_{D5})
\end{equation}
where $L_{class}$ is the sum of the losses of each classification subnet module, $L_{box}$ is the sum of the losses of each regression subnet module. Their sum represent the standard RetinaNet loss. $L_{D3},L_{D4},L_{D5}$ are the losses of each discriminator module and each of them is given by $L_{D_i}=\frac{1}{2}(L_{D_{s,i}}+L_{D_{t,i}})$ where $L_{D_{s,i}}$ and $L_{D_{t,i}}$ are respectively the losses computed by the discriminators when receive in input respectively synthetic and real images and defined using the Focal loss~\cite{DBLP:journals/corr/abs-1708-02002}. $\lambda$ is the hyperparameter that balances RetinaNet and discriminators losses. 
\subsection{Domain adaptation through feature alignment and image to image translation}
\begin{figure}[t]
            \centering
            \includegraphics[width=1\textwidth]{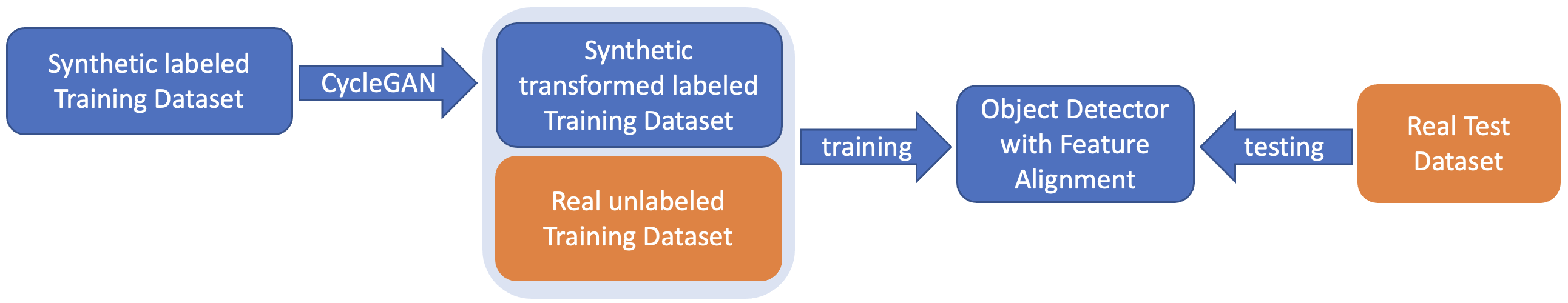}
            \caption{Pipeline used to combine feature alignment and image to image translation from synthetic to real techniques.}
            \label{fig:storf}
\end{figure}

\begin{figure}[t]
            \centering
            \includegraphics[width=1\textwidth]{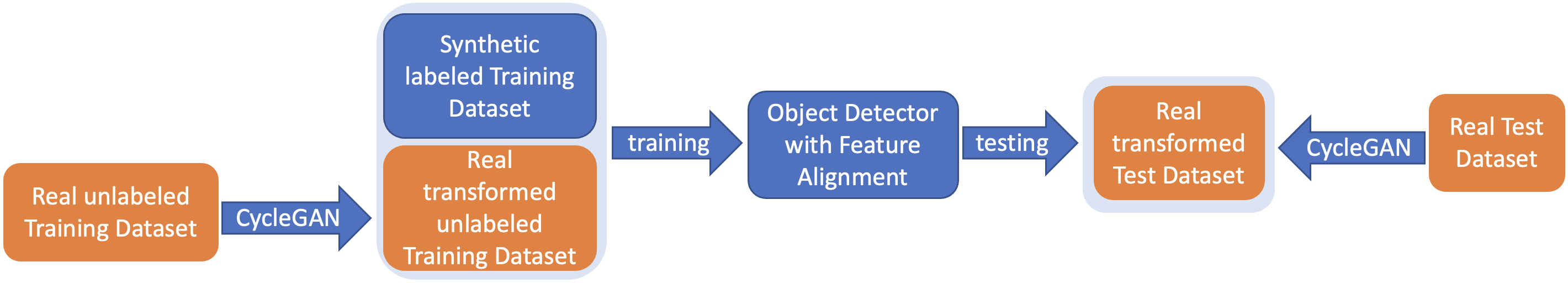}
            \caption{Pipeline used to combine feature alignment and image to image translation from real to synthetic techniques.}
            \label{fig:rtosf}
\end{figure}

We combine the feature alignment techniques presented in Section~\ref{featureAlignment}~and Section~\ref{DAretinanet} with image-to-image translation. This approach is similar to CyCADA proposed in~\cite{hoffman2017cycada} with the difference that we consider state-of-art feature alignment methods to perform the adaptation. We combine these techniques in two ways: 1) transforming synthetic labeled images to real, then training feature-alignment-based architectures using transformed labeled and real unlabeled images (Figure~\ref{fig:storf}); 2) transforming real unlabeled images to synthetic, then training feature alignment based architecture using synthetic labeled and transformed unlabeled images and testing on real images transformed to synthetic (Figure~\ref{fig:rtosf}).
\section{Experimental Settings and Results}
This section presents the proposed dataset, reports and analyze the results of the methods presented in the previous section and discusses the computational resources required to train all the models.
\label{result}
\subsection{Dataset}
\begin{figure*}[t!]
            \centering            
            \includegraphics[width=.2\textwidth]{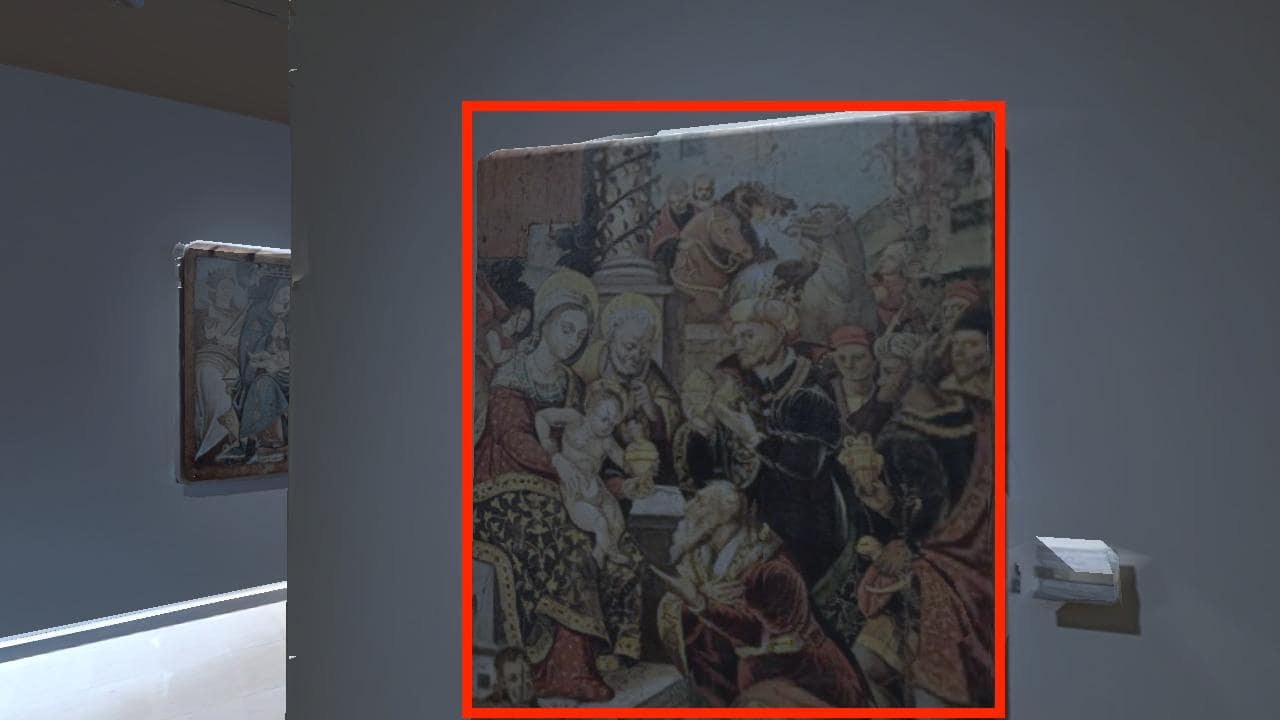}
            \includegraphics[width=.2\textwidth]{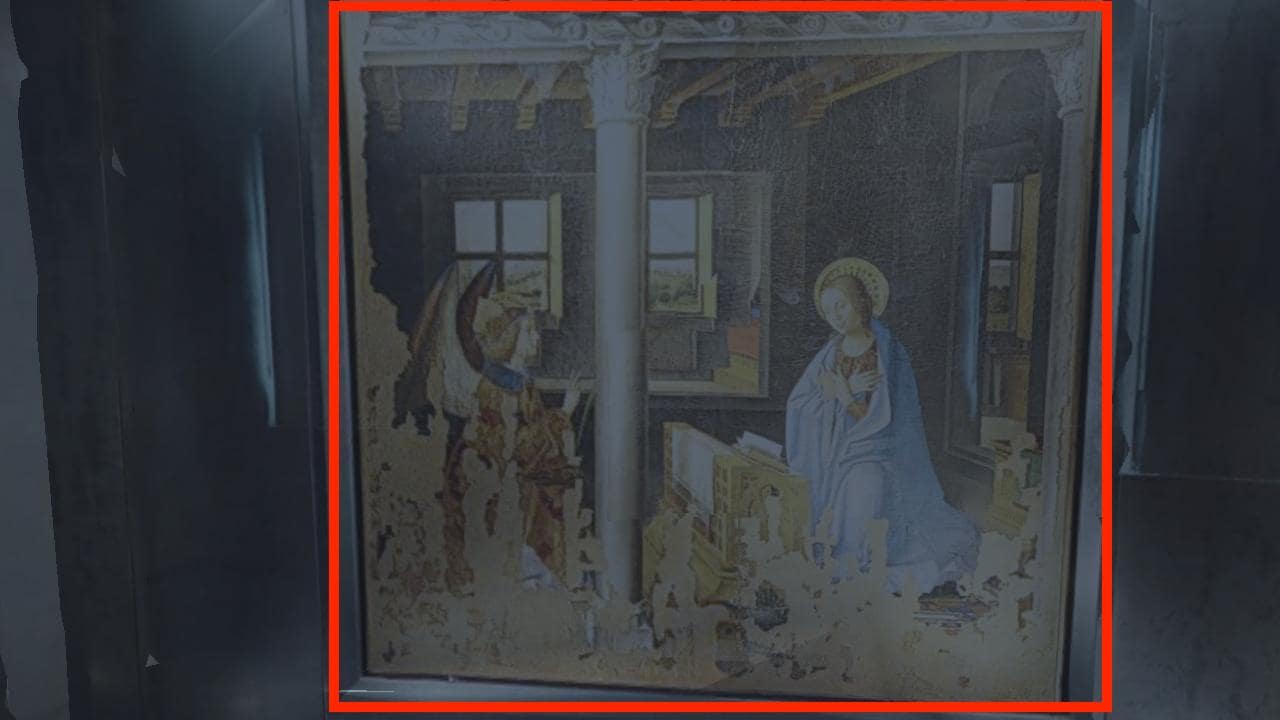}
            \vspace{0.1cm}
            \includegraphics[width=.2\textwidth]{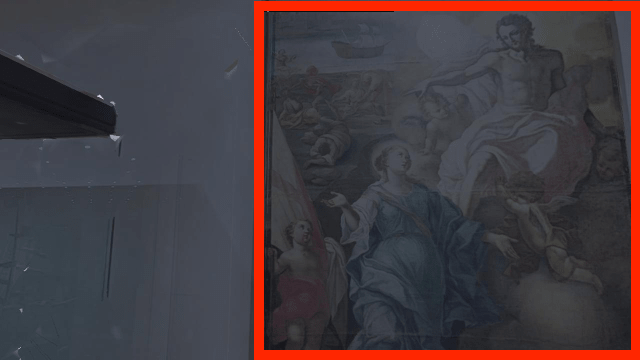}
            \includegraphics[width=.2\textwidth]{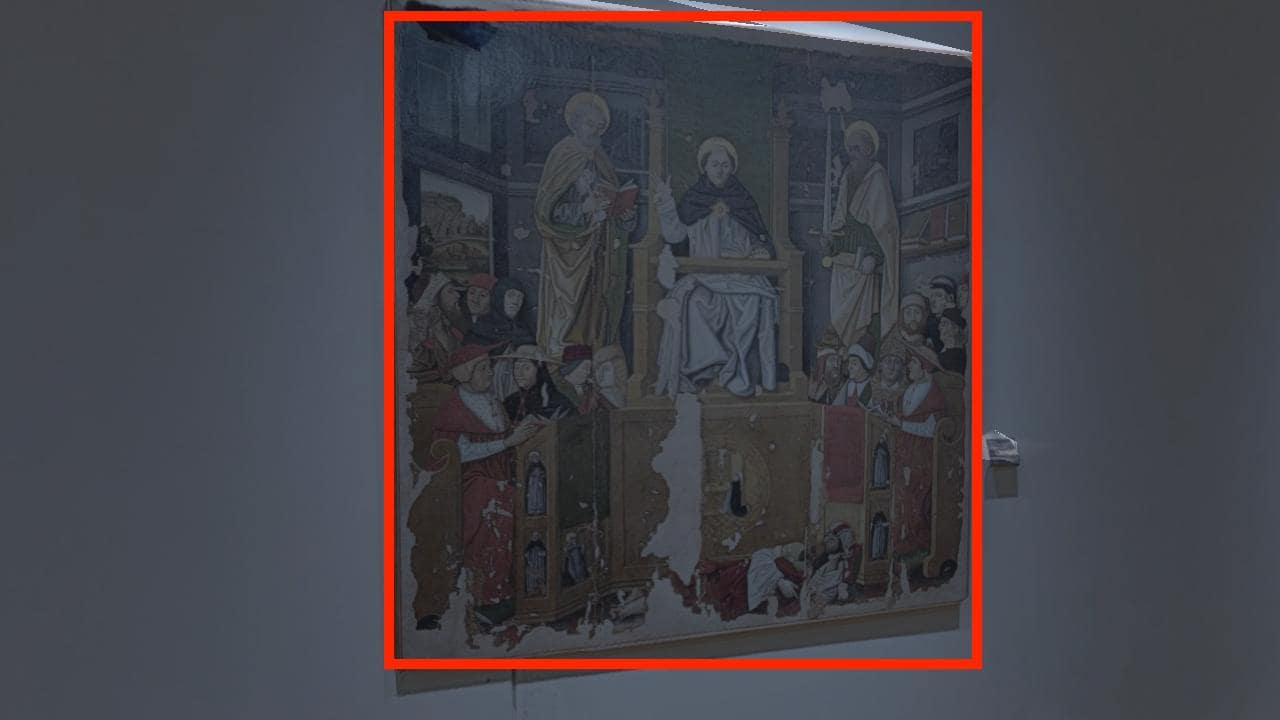}
            \includegraphics[width=.2\textwidth]{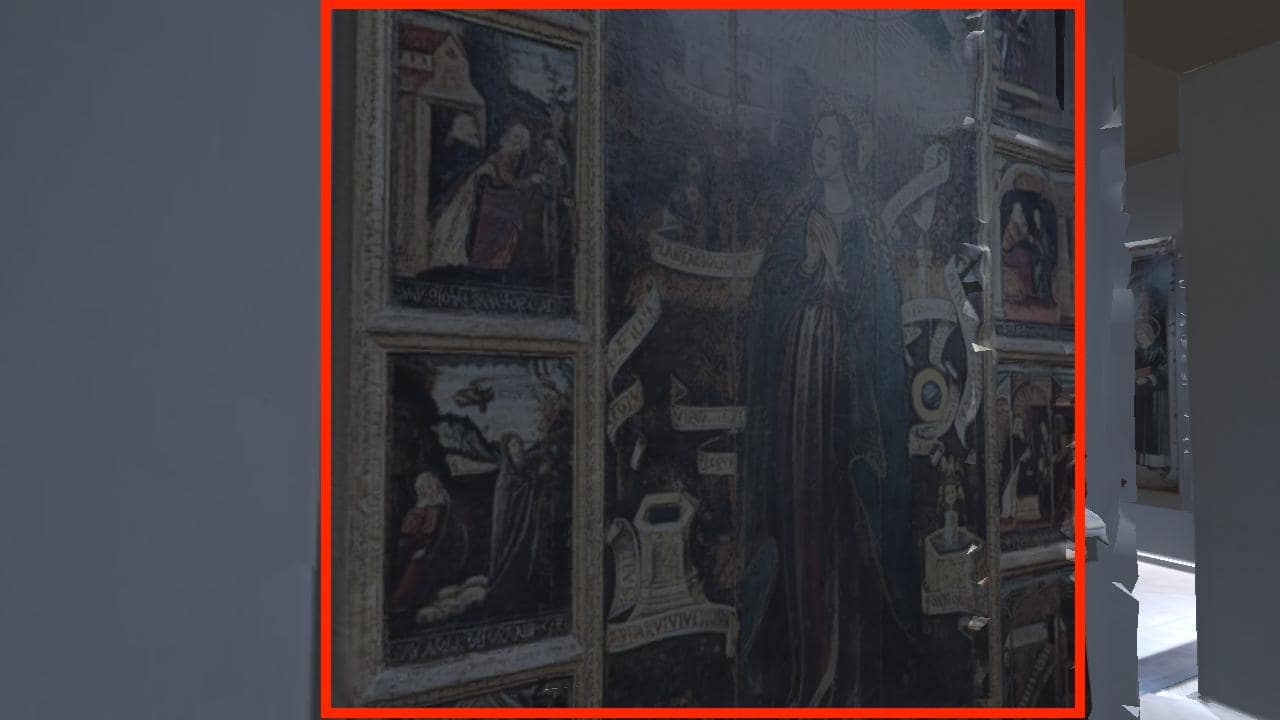}
            \includegraphics[width=.2\textwidth]{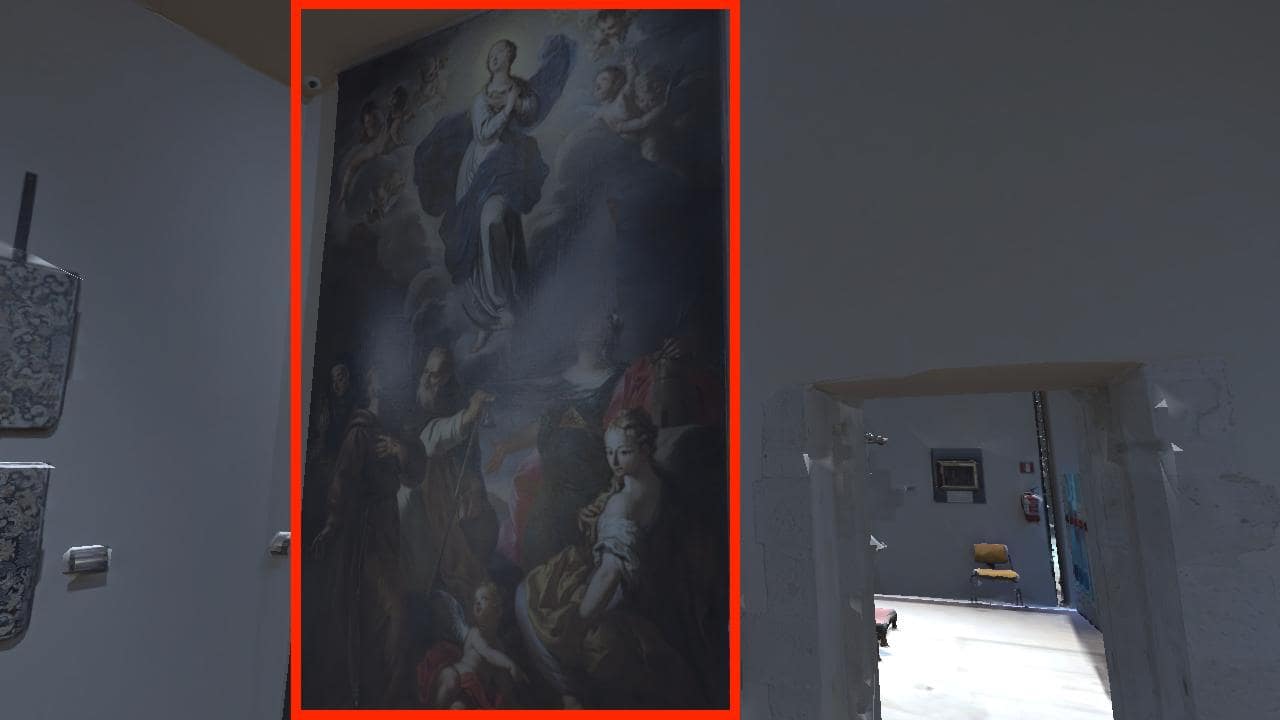}
            \vspace{0.1cm}
            \includegraphics[width=.2\textwidth]{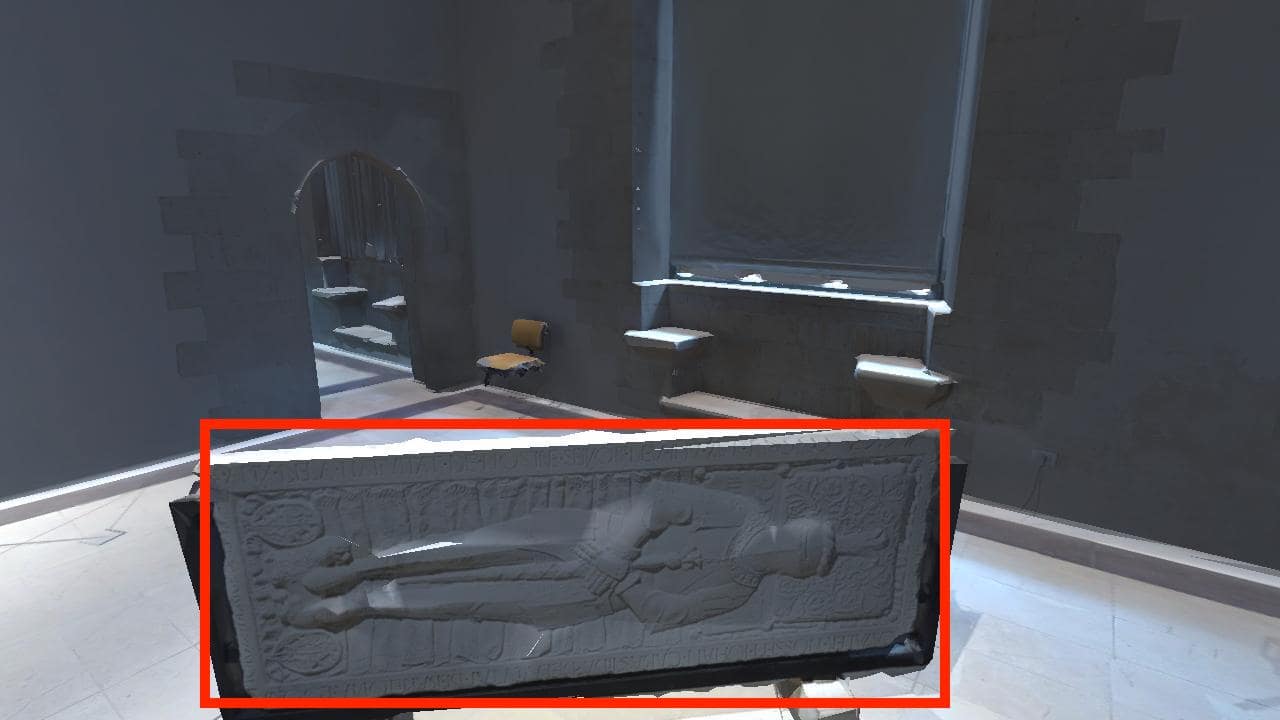}
            \includegraphics[width=.2\textwidth]{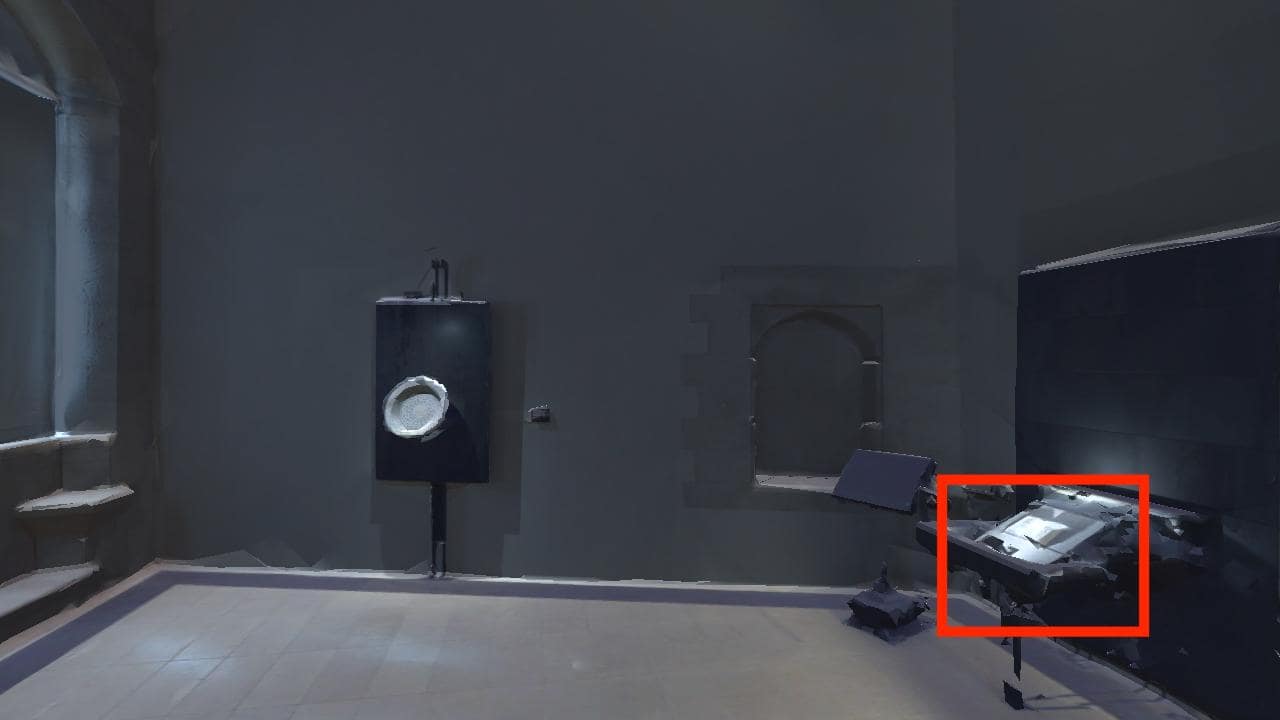}
            \includegraphics[width=.2\textwidth]{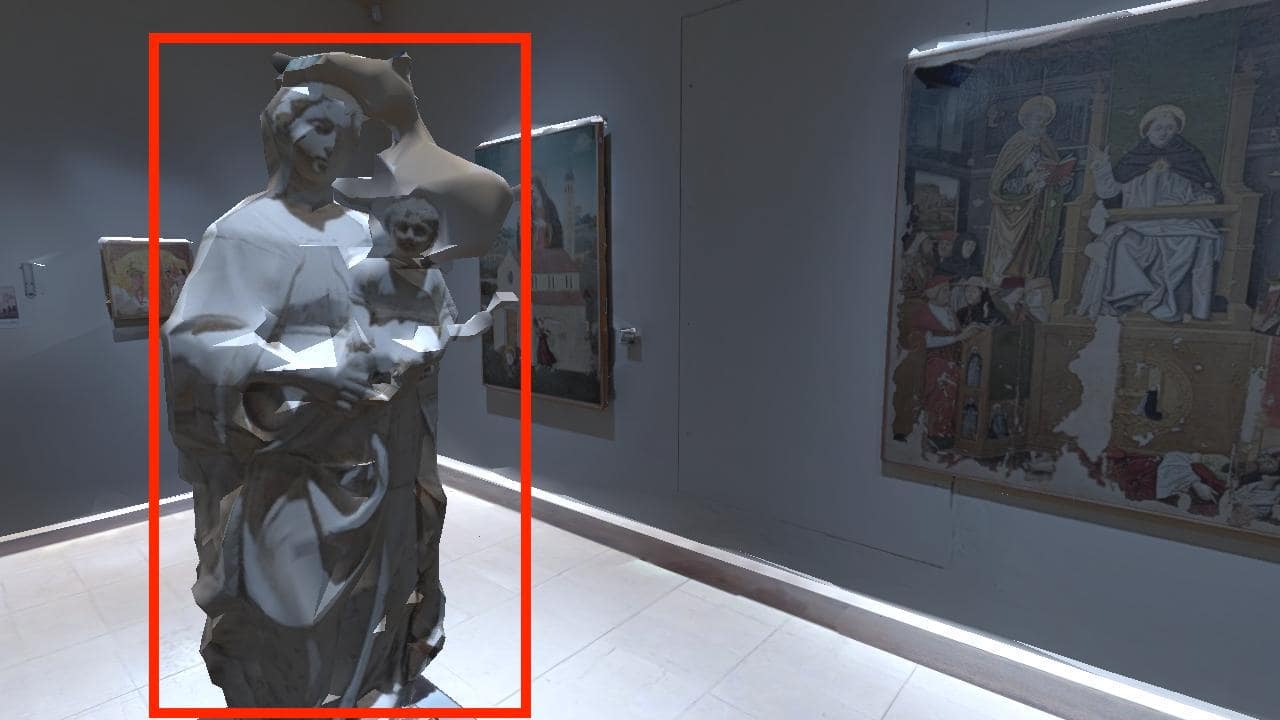}
            \includegraphics[width=.2\textwidth]{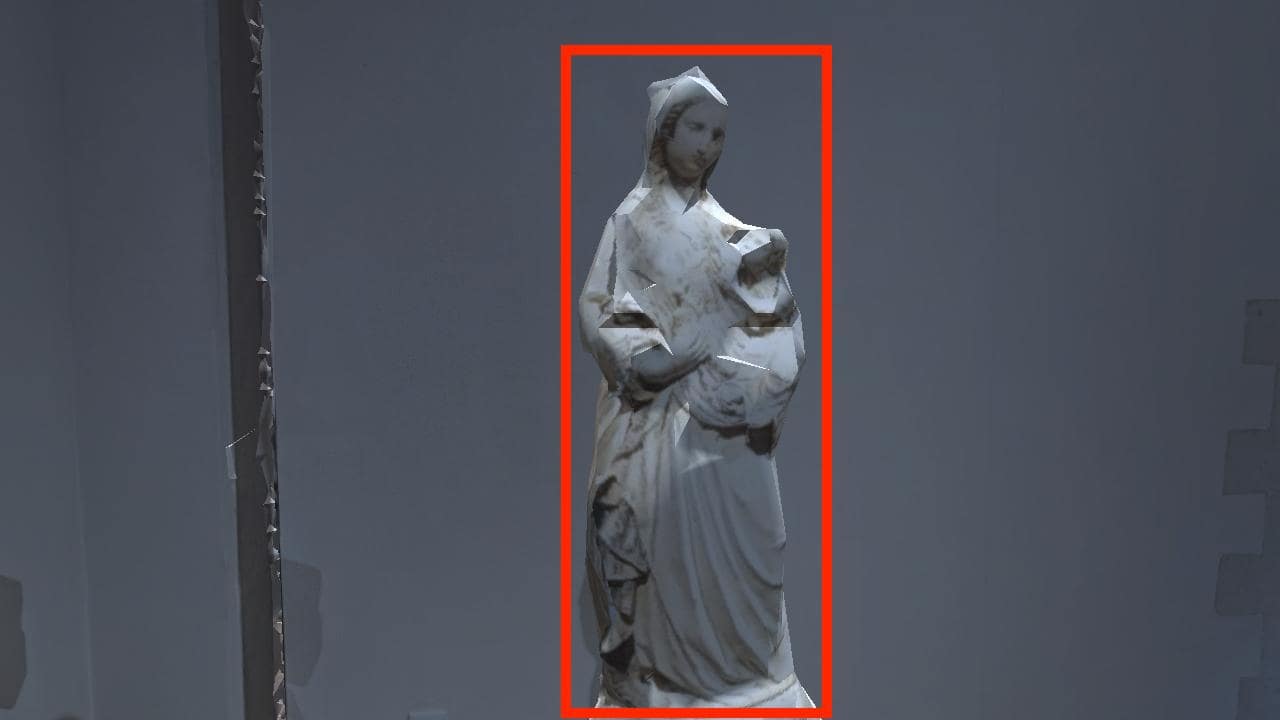}
            \includegraphics[width=.2\textwidth]{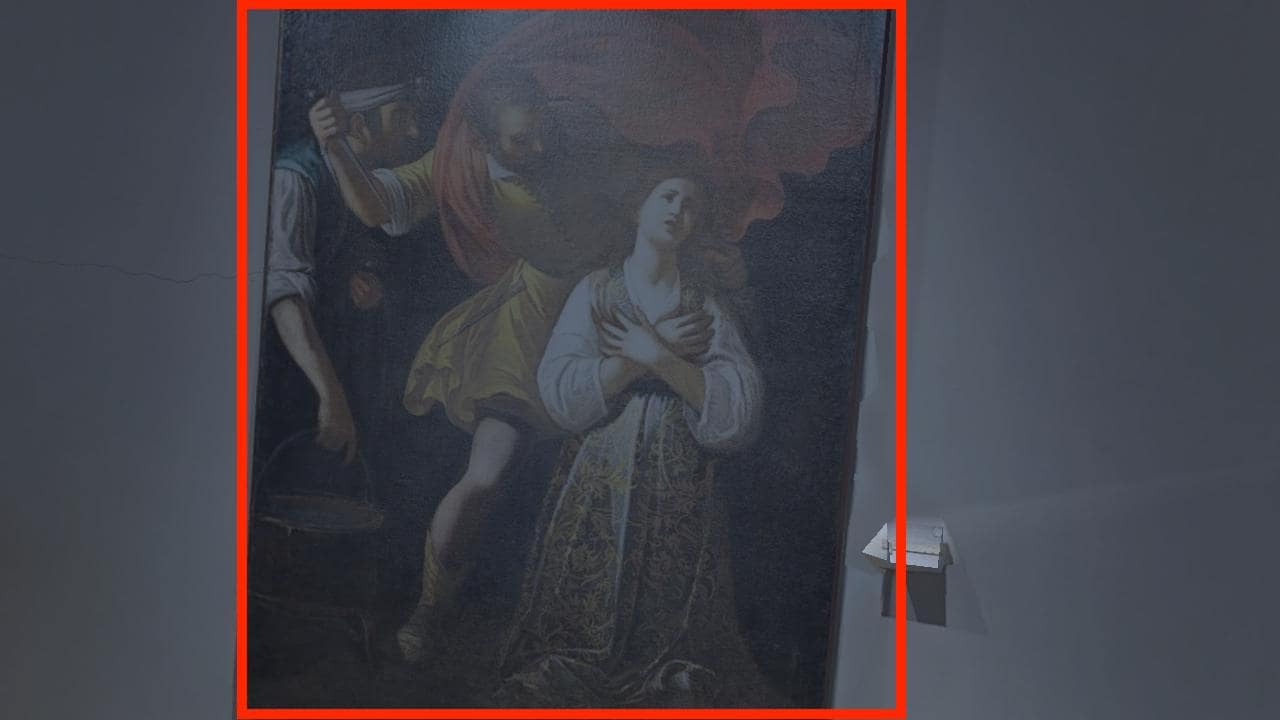}
            \includegraphics[width=.2\textwidth]{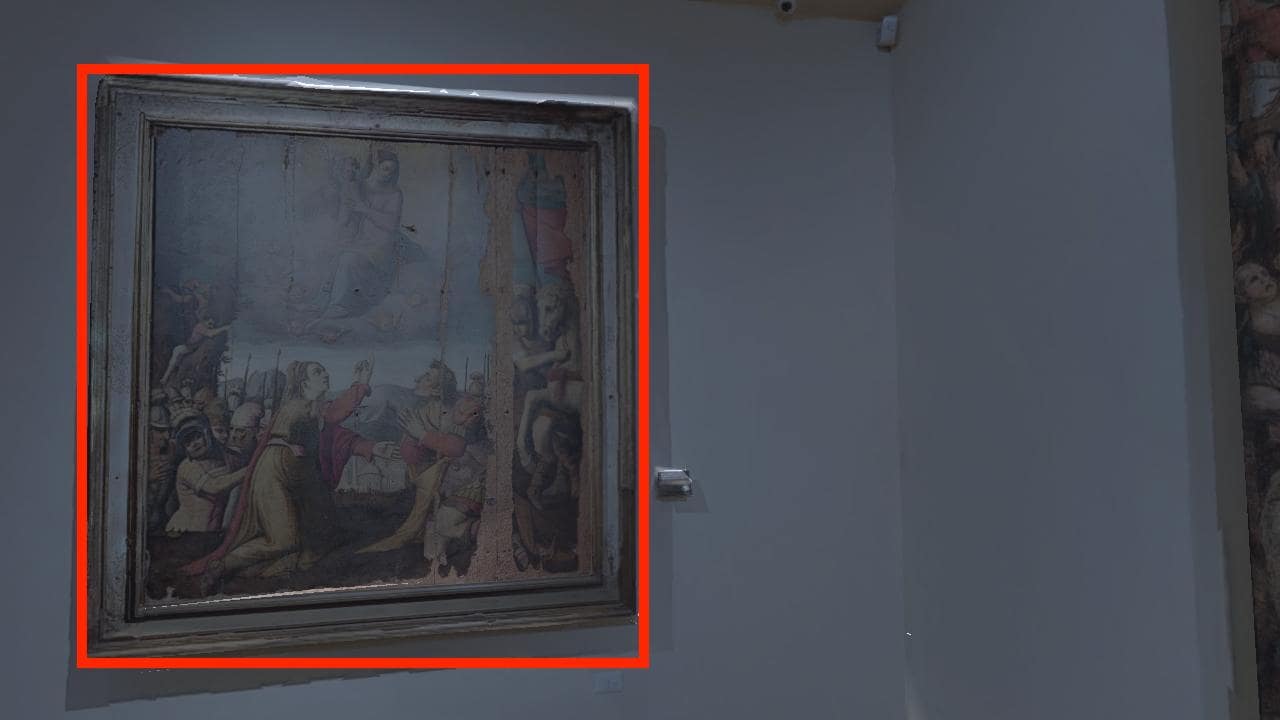}
            \includegraphics[width=.2\textwidth]{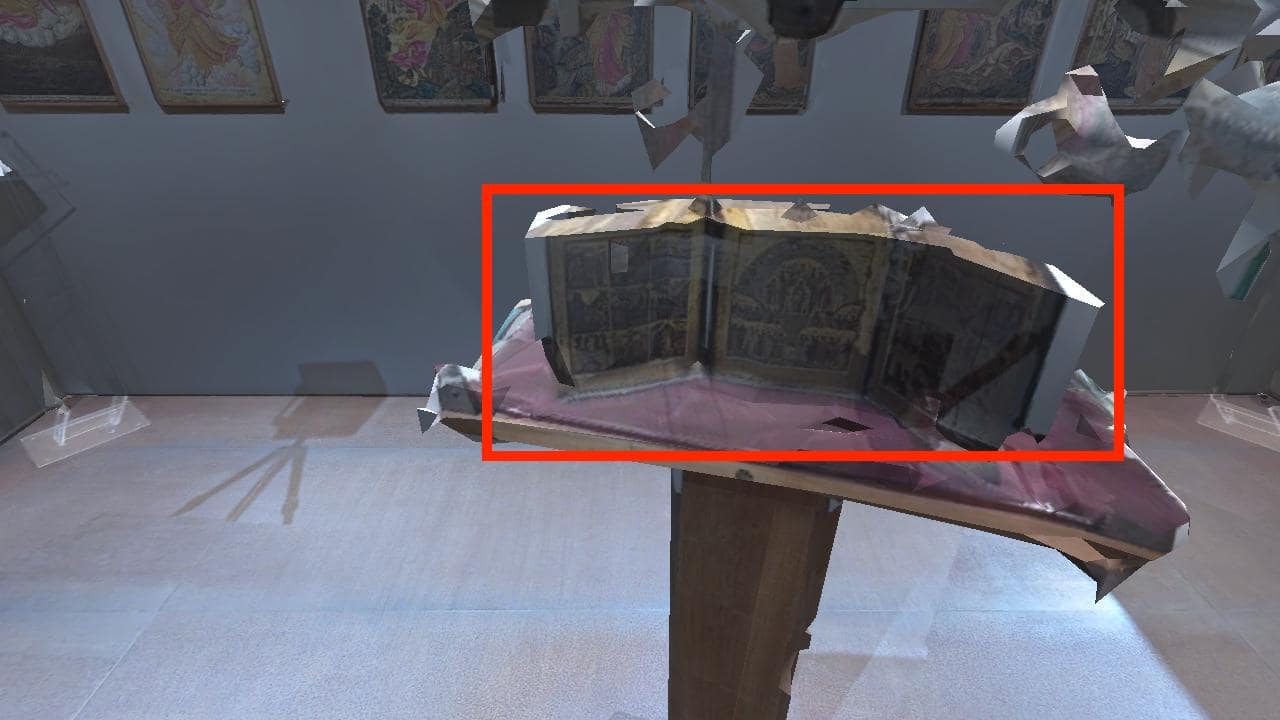}
            \includegraphics[width=.2\textwidth]{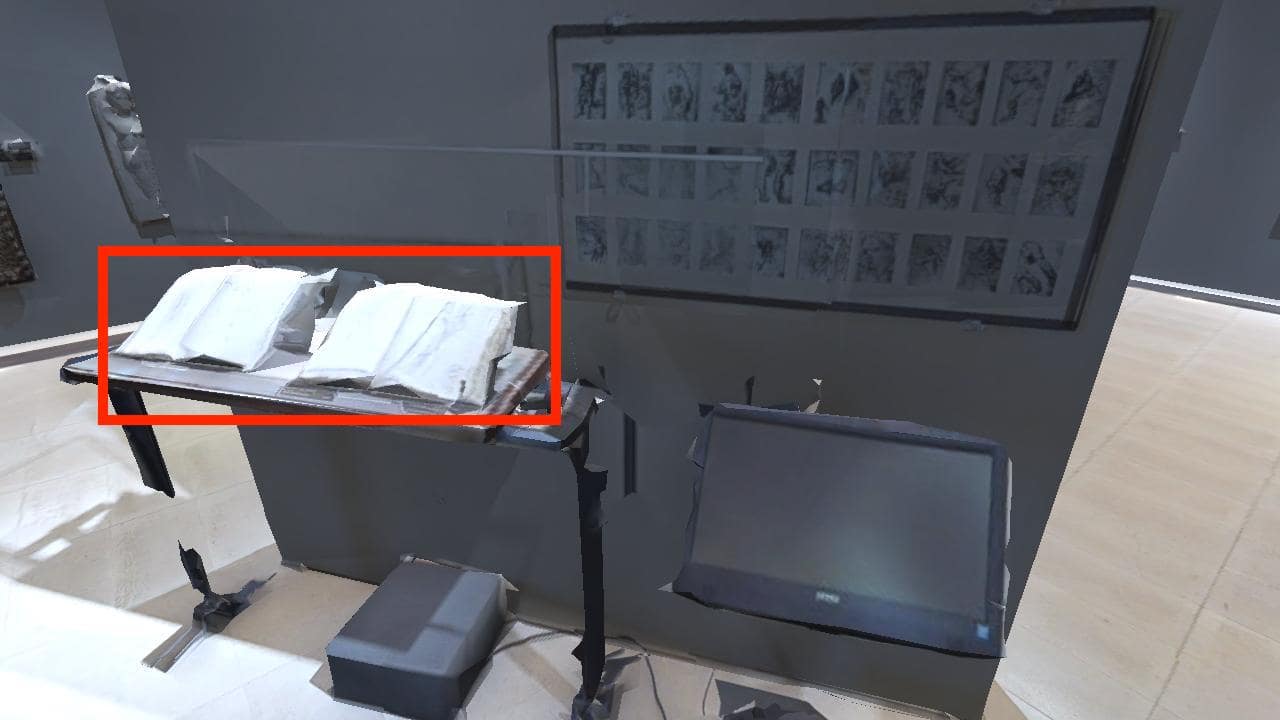}
            \includegraphics[width=.2\textwidth]{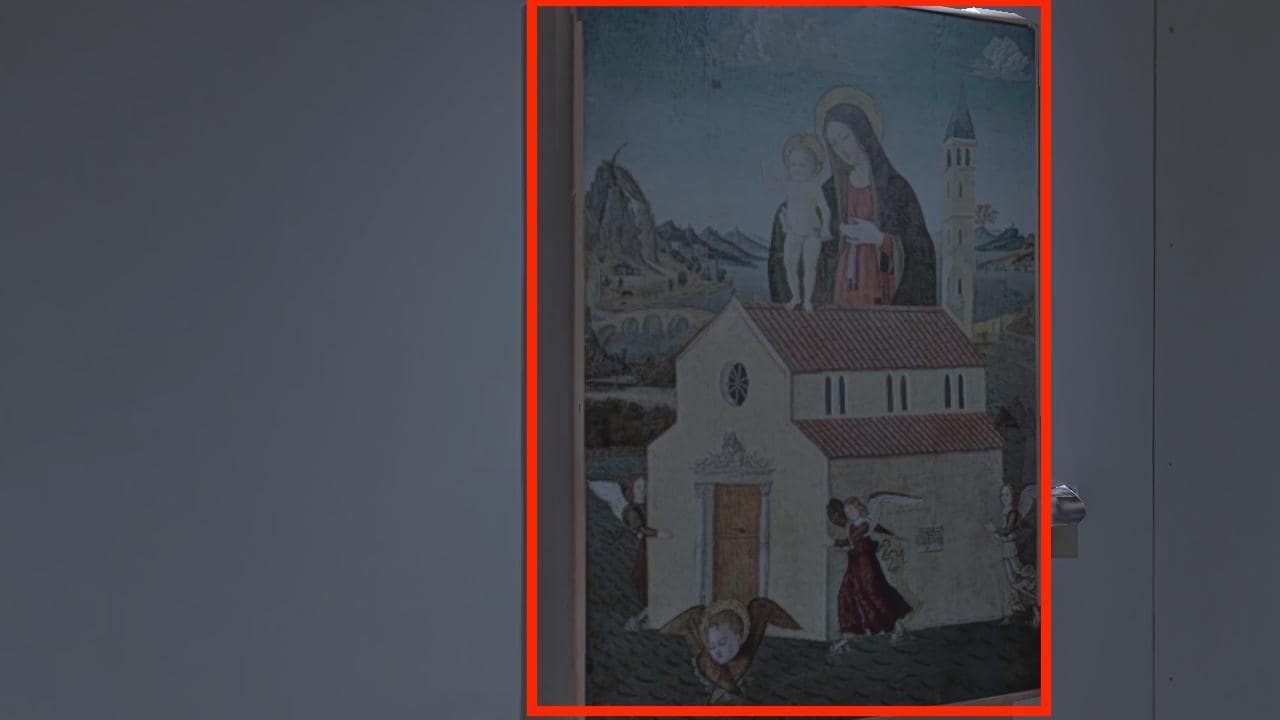}
            \includegraphics[width=.2\textwidth]{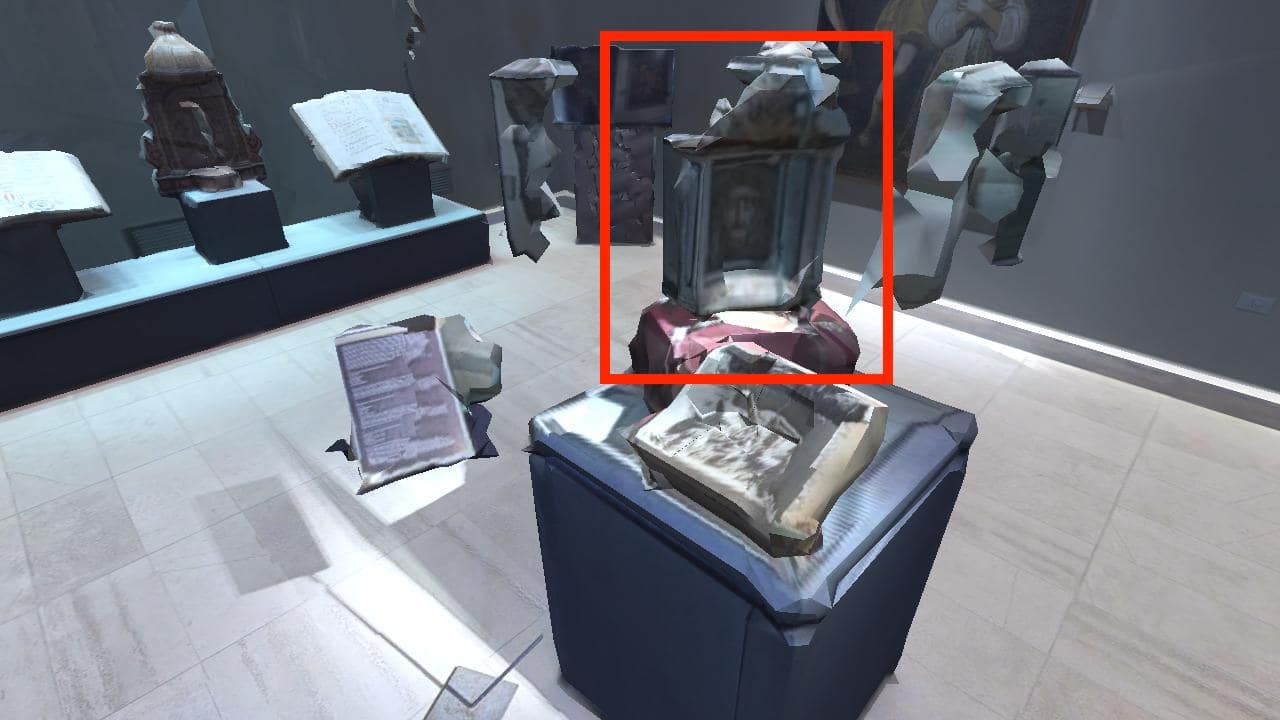}
            \caption{\label{fig:synthetic}Sample synthetic images of the 16 artworks of our dataset.}
            
            \vspace{0.4cm}
            \includegraphics[width=.2\textwidth]{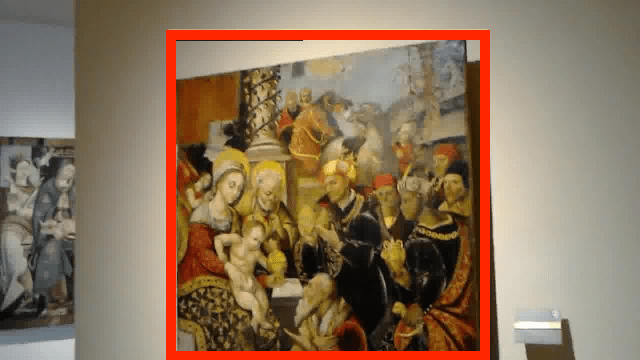}
            \includegraphics[width=.2\textwidth]{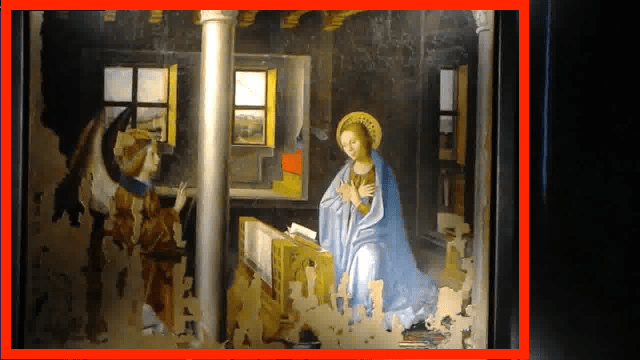}
            \vspace{0.1cm}
            \includegraphics[width=.2\textwidth]{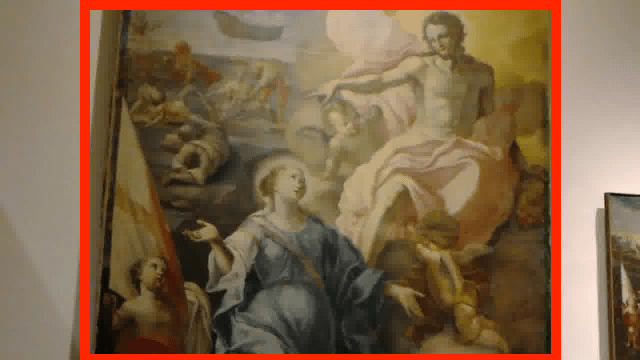}
            \includegraphics[width=.2\textwidth]{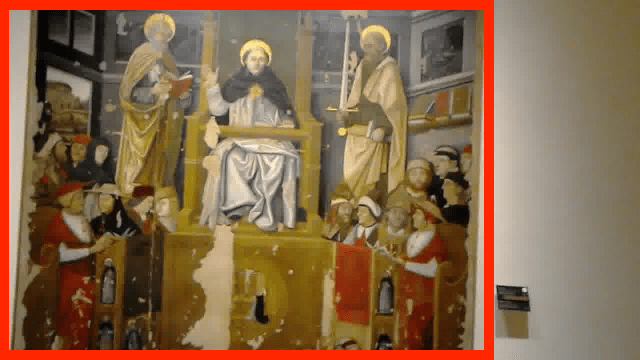}
            \includegraphics[width=.2\textwidth]{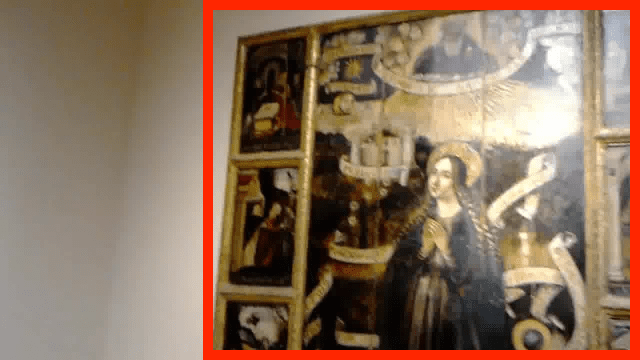}
            \includegraphics[width=.2\textwidth]{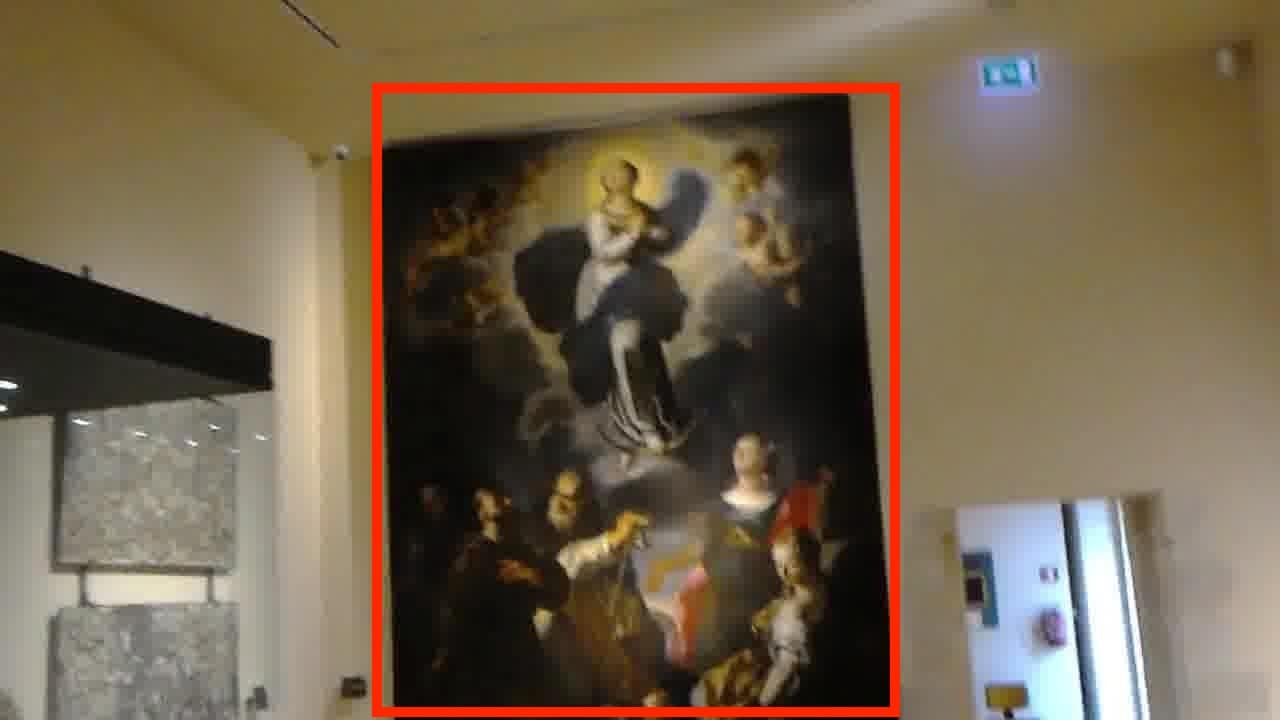}
            \includegraphics[width=.2\textwidth]{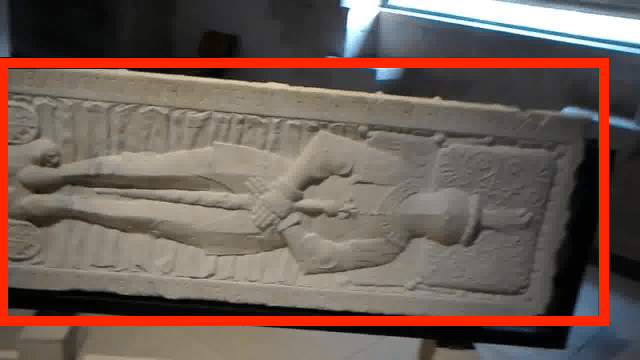}
            \vspace{0.1cm}
            \includegraphics[width=.2\textwidth]{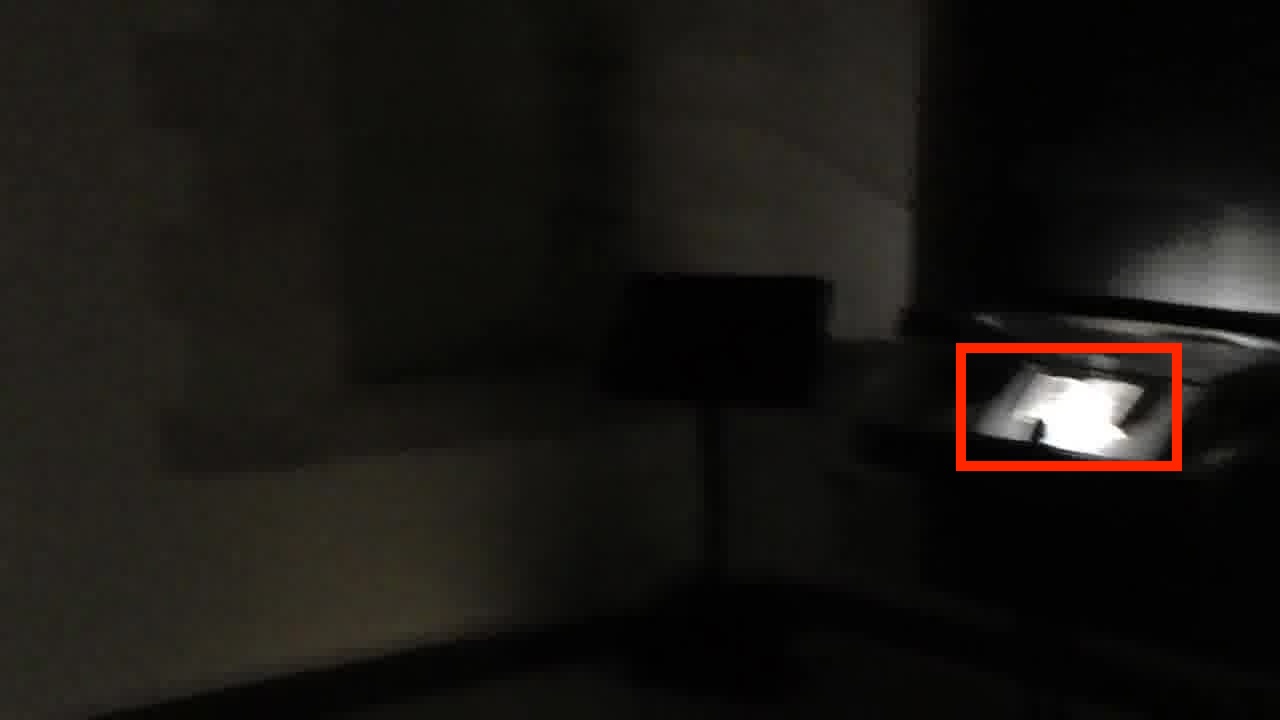}
            \includegraphics[width=.2\textwidth]{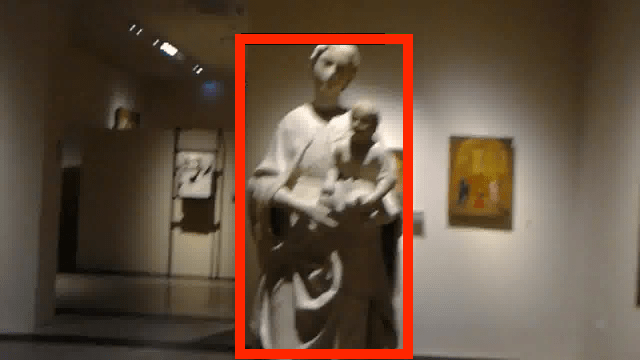}
            \includegraphics[width=.2\textwidth]{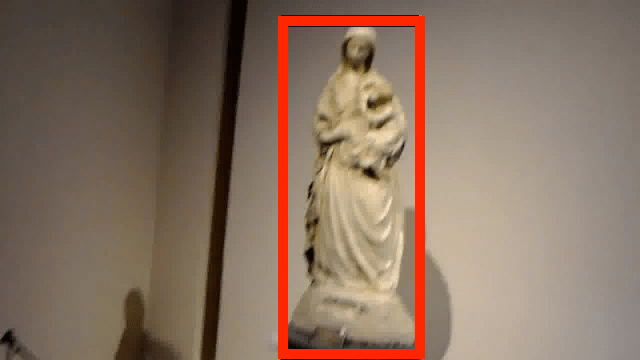}
            \includegraphics[width=.2\textwidth]{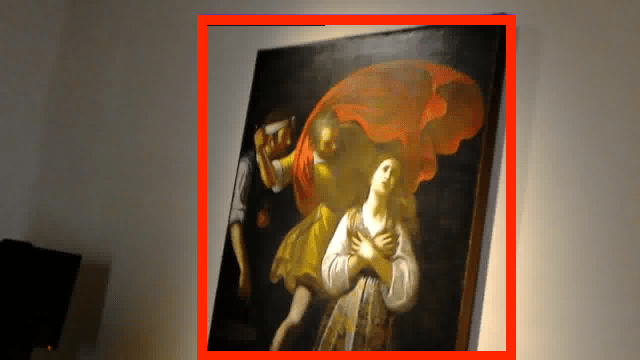}
            \includegraphics[width=.2\textwidth]{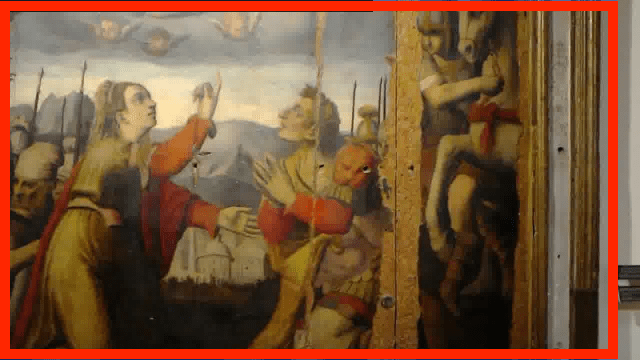}
            \includegraphics[width=.2\textwidth]{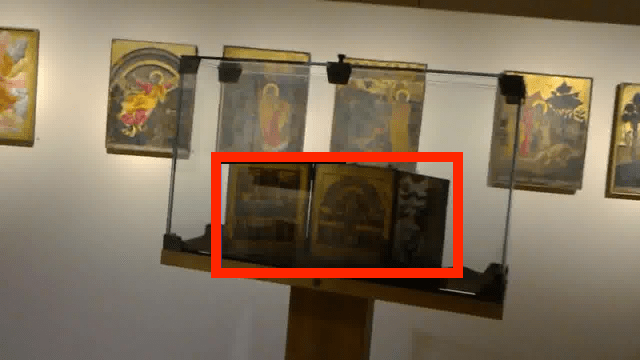}
            \includegraphics[width=.2\textwidth]{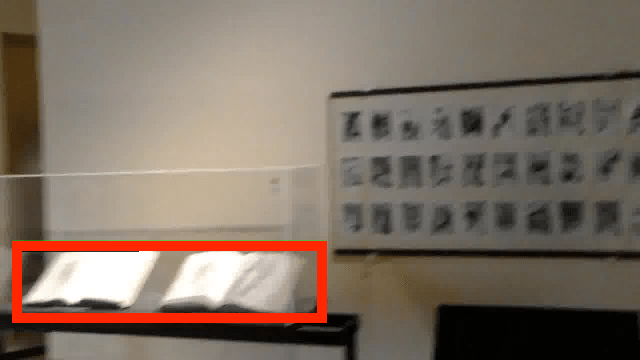}
            \includegraphics[width=.2\textwidth]{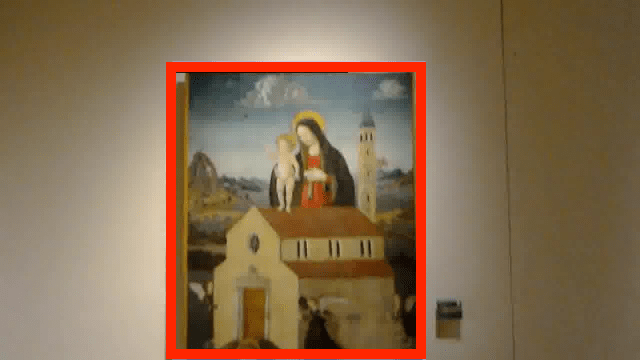}
            \includegraphics[width=.2\textwidth]{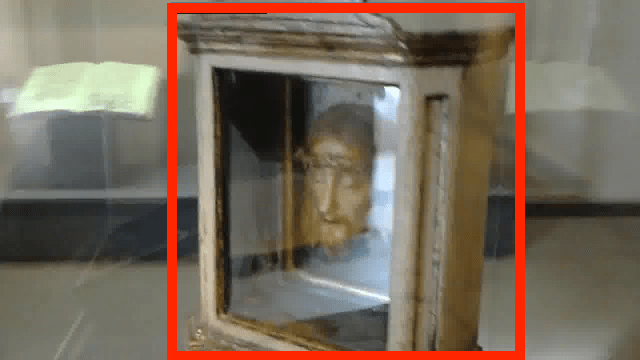}
            \caption{\label{fig:real}Sample real images of the 16 artworks of our dataset.}
\end{figure*}

The proposed dataset~\cite{pasqualino2020unsupervised} contains 16 objects that cover a variety of artworks which can be found in a museum like sculptures, paintings and books. Specifically, the dataset has been collected inside the cultural site ``Galleria Regionale di Palazzo Bellomo'' located in Siracusa, Italy\footnote{\url{http://www.regione.sicilia.it/beniculturali/palazzobellomo/}}. We generated 75244 synthetic labeled images, we have used a 3D model of the museum acquired using Matterport\footnote{https://matterport.com/}, of the 16 artworks using the public tool proposed by the authors of~\cite{orlando2020egocentric} (see Figure~\ref{fig:synthetic}). The tool proposed in~\cite{orlando2020egocentric} allows generate automatic labeled synthetic images, simulating a visitor who walks around the site while observing the artworks. Each image acquired during the simulation is associated to a semantic mask which allows to obtain bounding box annotations for each image. Real images of the same 16 artworks are taken from the EGO-CH dataset proposed in~\cite{Ragusa_2020} (see Figure~\ref{fig:real}), which contains videos of 70 subjects who visited two cultural sites which have been captured using a Microsoft HoloLens device. EGO-CH includes 176999 images manually annotated with bounding boxes. For the experiments, a subset of EGO-CH was taken into account. In particular, we considered 2190 images which contain the 16 artworks present in the synthetic dataset. To perform the experiments, we split both sets of synthetic and real images to training and a test set. We used 51284 synthetic and 1502 real images as training set and 23960 synthetic and 688 images as test set. The proposed dataset is available at the following URL: \url{https://iplab.dmi.unict.it/EGO-CH-OBJ-UDA/}.
\subsection{Experimental Settings}
We trained all the object detectors for 62K iterations starting from ImageNet~\cite{deng2009imagenet} pre-trained weights. We used Faster-RCNN and RetinaNet Detectron2~\cite{wu2019detectron2} architectures\footnote{\url{https://github.com/facebookresearch/detectron2}} with ResNet101~\cite{he2016deep} as backbone. The batch size has been set to 4 and the learning rate to 0.0002 for the first 30K iterations and multiplied by 0.1 for the remaining iterations. CycleGAN was trained for 60 epochs using the default parameters.  For DA-Faster RCNN\footnote{\url{https://github.com/krumo/Detectron-DA-Faster-RCNN}} and Strong-Weak\footnote{\url{https://github.com/VisionLearningGroup/DA_Detection}} we used the settings proposed by the authors in their respective works \cite{chen2018domain}~and~\cite{Saito_2019}. DA-RetinaNet\footnote{\url{https://github.com/fpv-iplab/DA-RetinaNet}} was implemented using Detectron2 and it was trained with a batch size of 6 and learning rate of 0.0002 for the first 30K iterations. Also in this case, the learning rate has been then multiplied by 0.1 for the remaining iterations.
\subsection{Baseline Results}
\begin{table}[t]
\caption{Performance of Faster RCNN and RetinaNet trained and tested on images from the same domain.}
\label{oracle}
\centering
\begin{tabular}{|c||c||c|}
\hline
 & \multicolumn{2}{c|}{mAP} \\
\hline
Model & Synthetic & Real\\
\hline
Faster RCNN & \textbf{93.08\%} & 92.04\%\\
\hline
RetinaNet & 91.67\% & \textbf{92.15\%}\\
\hline
\end{tabular}
\end{table}
\begin{table}[t]
\setlength{\tabcolsep}{1.5pt}
\caption{Performance of Faster RCNN and RetinaNet trained on synthetic images for a different amounts of iterations and tested on real images.}
\label{tab:naive}
\centering
\adjustbox{width=\linewidth}{
\begin{tabular}{|c||c||c||c||c||c||c||c|}
\hline
& \multicolumn{7}{c|}{Training Iterations} \\
\hline
Model & 6K & 12K & 22K & 32K & 42K & 52K & 62K \\
\hline
F. RCNN & 2.27\% & \textbf{9.67\%} & 5.79\% & 3.58\% & 3.33\% & 3.81\% & 3.62\% \\ \hline
RetinaNet & 9.83\% & \textbf{14.44\%} & 13.22\% & 12.31\% & 12.09\% & 12.44\% & 11.97\% \\
\hline
\end{tabular}
}
\end{table}
Table~\ref{oracle} reports the results of the two models when they are trained and tested on the same domain. As can be noted, when images came from the same distribution, these algorithms achieve good performance. Table~\ref{tab:naive} shows the performance achieved by Faster RCNN and RetinaNet when trained on synthetic images and tested on real images. The results highlight that models trained for few iterations generalize better than models trained for more iterations. RetinaNet is in general more robust to domain shift than Faster RCNN. In particular, RetinaNet trained for 12K iterations achieves an mAP of 14.44\% vs 9.67\% obtained by Faster RCNN and 11.97\% vs 3.62\% considering 62K iterations. This suggests that training for more iterations both models increases the domain gap between the two distributions because the models learn to extract features specific to the source domain that do not generalize to the target domain. It is worth noting that, even the best RetinaNet model (14.44\%) exhibits a drastic drop in performances if compared with the results of Table~\ref{oracle} (92.15\%). This is due to the domain shift between synthetic images used for training and real images used for test.
\subsection{Image-to-Image translation Results}
\begin{figure*}[t]
 \centering
    \includegraphics[width=0.47\linewidth]{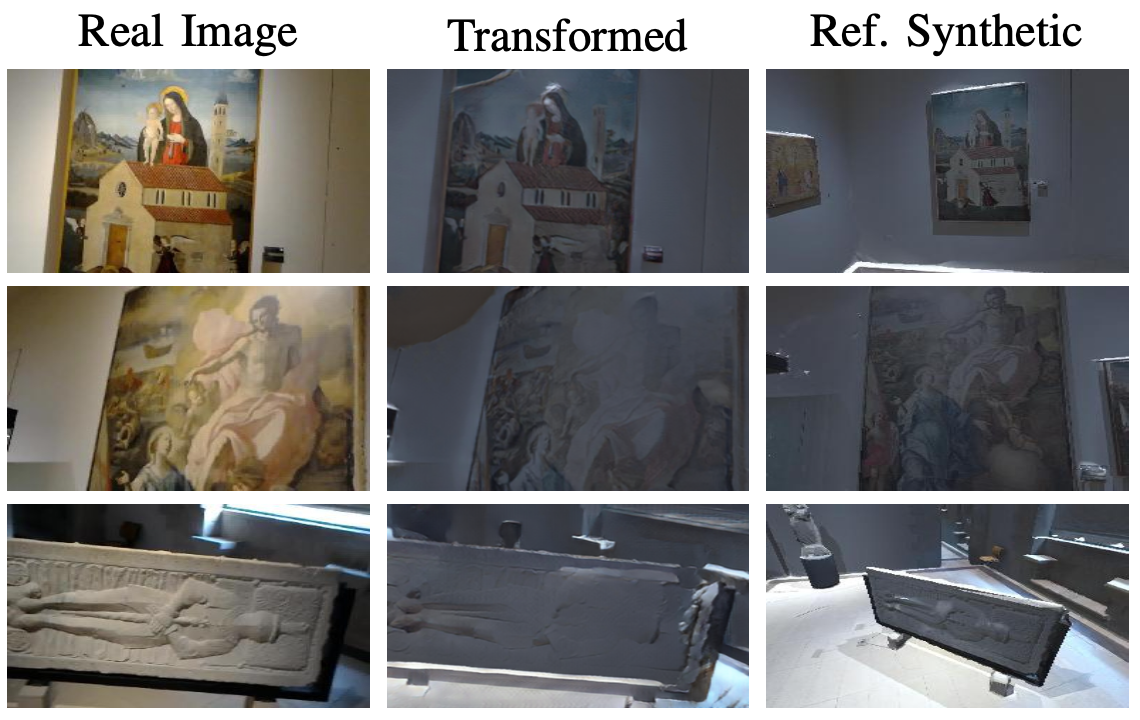}
    \quad
    \includegraphics[width=0.47\linewidth]{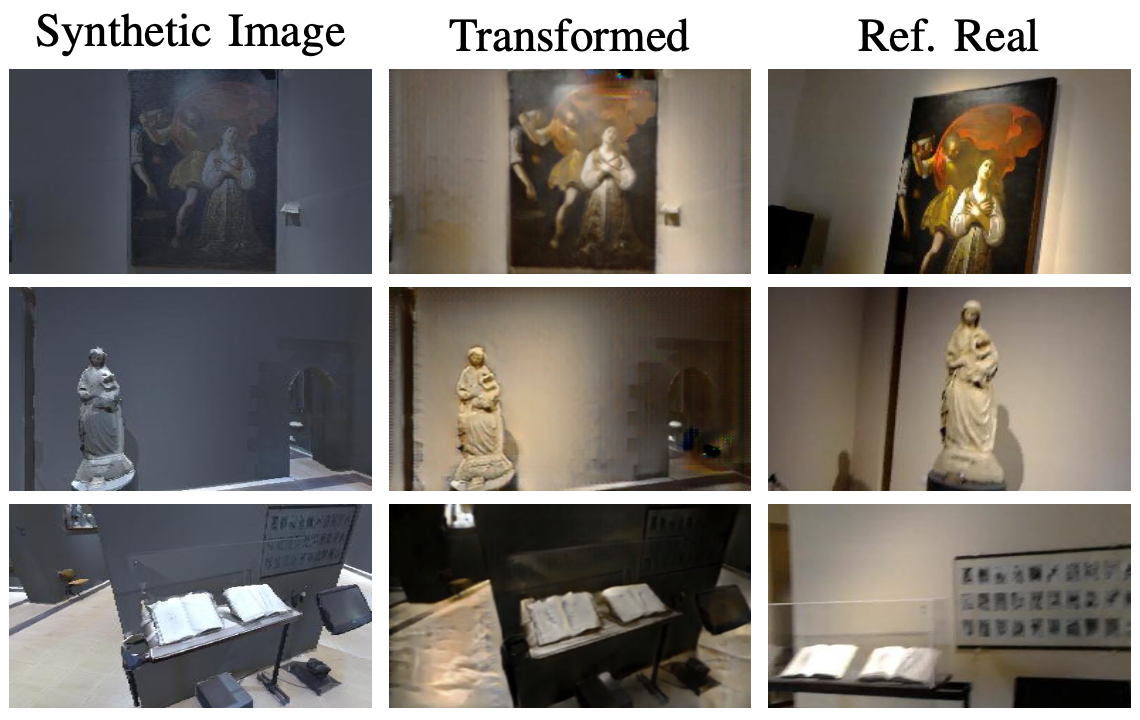}
    \caption{Qualitative CycleGAN results. We show the source domain (real synthetic), the transformed image, and a reference image for visual comparison. Left: transformation from real to synthetic. Right: transformation from synthetic to real.}
    \label{fig:qualitativeresultcyclegan}
\end{figure*}

\begin{table}[t!]
\setlength{\tabcolsep}{1.5pt}
\caption{Results obtained transforming real images to synthetic at test time. The models have been trained on synthetic images. N.A. stands for No Adaptation.}
\label{tab:RtoS}
\centering
\adjustbox{width=\linewidth}{
\begin{tabular}{|c||c||c||c||c||c||c||c|}
\hline
 & & \multicolumn{6}{c|}{Training epochs for CycleGAN} \\
\hline
Model (iter) & N.A. & 10 & 20 & 30 & 40 & 50 & 60 \\
\hline
F. RCNN (62K) & 3.62\% & 25.16\% & 25.49\% & 25.51\% & 26.68\% & 27.65\% & \textbf{28.25\%} \\ \hline
RetinaNet (62K) & 11.97\% & 27.30\% & 32.14\% & \textbf{34.15\%} & 32.66\% & 32.79\% & 32.82\% \\
\hline

F. RCNN (12K) & 9.67\% & 29.93\% & 32.84\% & 33.95\% & 31.45\% & \textbf{34.19\%} & 31.58\% \\ \hline
RetinaNet (12K) & 14.44\% & 34.51\% & 35.45\% & 34.84\% & 35.34\% & \textbf{35.76\%} & 35.74\% \\

\hline

\end{tabular}
}
\end{table}
\begin{table}[t!]
\setlength{\tabcolsep}{1.5pt}
\caption{Results obtained training the models on synthetic images transformed to real and tested on real images. N.A. stands for No Adaptation.}
\label{StoR}
\centering
\adjustbox{width=\linewidth}{
\begin{tabular}{|c||c||c||c||c||c||c||c|}
\hline
&&\multicolumn{6}{c|}{Training epochs for CycleGAN} \\
\hline
Model & N.A. & 10 & 20 & 30 & 40 & 50 & 60 \\
\hline
F. RCNN & 9.67\% & 18.76\% & 20.92\% & 21.22\% & 23.17\% & 24.45\% & \textbf{26.03\%} \\ \hline
RetinaNet & 14.44\% & 40.13\% & 44.29\% & 46.05\% & 47.89\% & 49.96\% & \textbf{55.54\%} \\ 
\hline
\end{tabular}
}
\end{table}
Table~\ref{tab:RtoS} shows the results of Faster RCNN and RetinaNet tested on real images transformed to synthetic using CycleGAN. We analyzed the performances of both models trained for 12K and 62K iterations to explore the impact of overfitting. As shown in the Table~\ref{tab:RtoS}, CycleGAN improves the performance of both models. RetinaNet performs better than Faster RCNN. Indeed Faster RCNN achieves performance similar to RetinaNet only when tested on images transformed using a CycleGAN model trained for 50 epochs. Table~\ref{StoR} reports the results of both models trained using synthetic images transformed to real. In this case, the performance of Faster RCNN (26.03\%) are lower than the previous method which uses images translated from real to synthetic (28.25\%). RetinaNet increases its performance by $\sim$20\% from 35.76\% to 55.54\%. Even in this case, the results seem to confirm that RetinaNet is more robust to domain shift. While training CycleGAN for more epochs may allow for minor improvements, it should be noted that training CycleGAN for 60 epochs required about 61 days.
A detailed discussion on the training times of all methods is provided in Section~\ref{computational_re}. Figure~\ref{fig:qualitativeresultcyclegan} shows qualitative example obtained translating images from real to synthetic and vice versa. The first row of Figure~\ref{fig:qualitativeresultcyclegan} (a) shows an example of successful translation. In the second row the image is not correctly transformed due to light reflection, whereas in the third row the texture is destroyed during the transformation. First two rows of Figure~\ref{fig:qualitativeresultcyclegan} (b) show an example of successful translation while the last rows show a bad translation example where the background contains many artifacts.
\subsection{Feature Alignment and Image-to-Image translation Results}
Table~\ref{tab:feature_Align} reports the results of the methods based only on feature alignment and combined with CycleGAN. As can be seen from Table~\ref{tab:feature_Align}, the proposed DA-RetinaNet achieves better performances when compared to other methods. Without image-to-image translation, DA-RetinaNet obtains an mAP of 31.04\% which is an increase in performance of about 6\% as compared to Strong-Weak (25.12\%). The improvement is about 11\% when the models are combined with SynToReal CycleGAN (58.01\% vs 47.70\%). Furthermore, it is worth noting that all models benefit from a performance improvement which varies between 21\% and 27\% when combined with CycleGAN.
\begin{table}[t!]
\caption{Results of DA-Faster RCNN, Strong-Weak and the proposed DA-RetinaNet combined  with two different image-to-image translation approaches.}
\label{tab:feature_Align}
\centering
\begin{tabular}{|c||c||c||c|}
\hline
& \multicolumn{3}{c|}{image-to-image translation} \\
\hline
Model & None & Real2Syn & Syn2Real\\
\hline
DA-Faster RCNN & 12.94\% & 19.88\% & 33.20\%\\
\hline
Strong-Weak & 25.12\% & 33.33\% & 47.70\%\\
\hline
DA-RetinaNet & \textbf{31.04\%} & \textbf{37.49\%} & \textbf{58.01\%}\\
\hline
\end{tabular}
\end{table}
\subsection{Ablation Study}
Table~\ref{tab:ablation} reports the results of DA-RetinaNet considering one, two or three Discriminators $D_i$ without any image-to-image translation technique. As shown in the table, aligning features using the paradigm of adversarial learning improves in each case the performance of standard RetinaNet. $D_3$, the discriminator that aligns low level features, doubles the performance with respect to the standard RetinaNet model achieving a mAP of 28.61\% vs 14.44\%. The use of $D_4$ and $D_5$ allows to achieve similar performance (16.38\% and 15.84\%) improving the baseline results by about 1.5\%. This is probably due to the design of the RetinaNet architecture. Indeed between the feature map $C_4$ and $C_5$ there are few convolutive layers. Combining the two discriminators which achieve the best performance allows to improve the mAP of about 2\% (28.61\% vs 30.52\%). The best model is obtained by using all the discriminators (31.04\%), which is our suggested design, as shown in Figure~\ref{fig:DA-Retinanet}
\begin{table}[t!]
\caption{Ablation study about the impact of each discriminator $D_i$.}
\label{tab:ablation}
\centering
\begin{tabular}{|c||c||c||c|c|}
\hline
Model & $D_3$ & $D_4$ & $D_5$ & mAP\\
\hline
RetinaNet (62K) &  &  &  & 11.97\%\\
\hline
RetinaNet (12K) &  &  &  & 14.44\%\\
\hline
DA-RetinaNet &  &  & \checkmark & 15.84\%\\
\hline
DA-RetinaNet &  & \checkmark & & 16.38\%\\
\hline
DA-RetinaNet & \checkmark &  &  & 28.61\%\\
\hline
DA-RetinaNet & \checkmark & \checkmark &  & 30.52\%\\
\hline
DA-RetinaNet & \checkmark & \checkmark & \checkmark & \textbf{31.04\%}\\
\hline
\end{tabular}
\end{table}
\subsection{Qualitative Results and Summary table}
\begin{table}[t!]
\caption{Summary table of the analyzed methods.}
\label{tab:summary}
\centering
\begin{tabular}{|c||c||c|}
\hline
Object Detector & Adaptation & mAP\\
\hline
Faster RCNN & None & $9.67\%$ \\
\hline
RetinaNet & None & $14.44\%$ \\
\hline
Faster RCNN & Real2Syn (Test set) & $34.19\%$ \\ 
\hline
RetinaNet & Real2Syn (Test set) & $35.76\%$ \\
\hline
Faster RCNN & Syn2Real (labeled Training set) & $26.03\%$ \\
\hline
RetinaNet & Syn2Real (labeled Training set) & $55.54\%$ \\
\hline
DA-Faster RCNN & Feat.Align. & $12.94\%$ \\
\hline
DA-Faster RCNN &
\shortstack{Feat.Align.+Real2Syn \\ (Test set and unlabeled Training set)}& $19.88\%$ \\
\hline
DA-Faster RCNN &\shortstack{Feat.Align.+Syn2Real \\ (labeled Training set)} & $33.20\%$ \\
\hline
Strong-Weak & Feat.Align. & $25.12\%$ \\
\hline
Strong-Weak & \shortstack{Feat.Align.+Real2Syn \\ (Test set and unlabeled Training set)} & $33.33\%$ \\
\hline
Strong-Weak & \shortstack{Feat.Align.+Syn2Real \\ (labeled Training set)} & $47.70\%$ \\
\hline
DA-RetinaNet & Feat.Align. & $31.04\%$ \\
\hline
DA-RetinaNet & \shortstack{Feat.Align.+Real2Syn \\ (Test set and unlabeled Training set)} & $37.49\%$ \\
\hline
DA-RetinaNet & \shortstack{Feat.Align.+Syn2Real \\ (labeled Training set)} & $\mathbf{58.01\%}$ \\
\hline
\end{tabular}
\end{table}
\begin{figure*}[t!]
            \centering
            \begin{minipage}{.19\textwidth}
            \centering
             Faster RCNN\\
            \end{minipage}
            \begin{minipage}{.19\textwidth}
            \centering
            DA-Faster RCNN\\
            \end{minipage}
            \begin{minipage}{.19\textwidth}
            \centering
             RetinaNet\\
            \end{minipage}
            \begin{minipage}{.19\textwidth}
            \centering
             Strong-Weak\\
            \end{minipage}
            \begin{minipage}{.19\textwidth}
            \centering
             DA-RetinaNet\\
            \end{minipage}
            
            \vspace{1mm}
            \includegraphics[width=.19\textwidth]{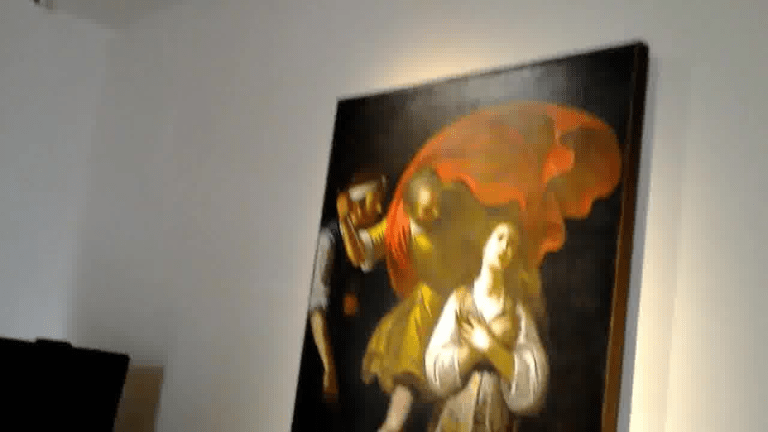}
            \includegraphics[width=.19\textwidth]{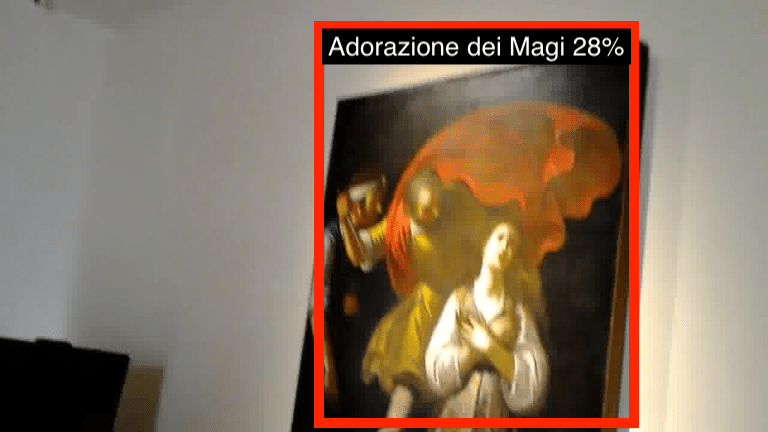}
            \includegraphics[width=.19\textwidth]{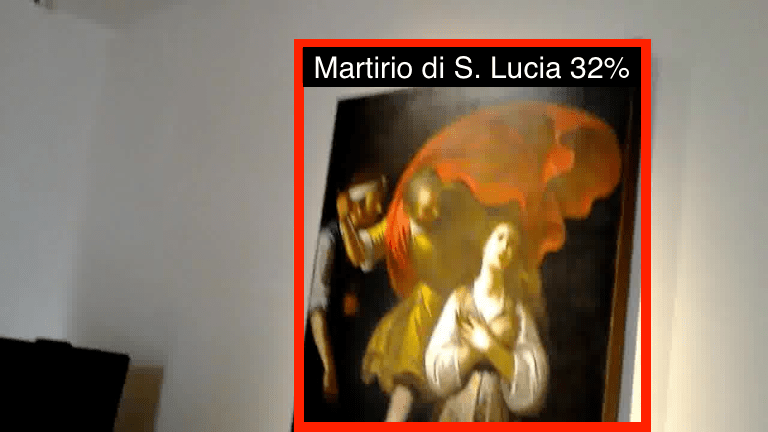}
            \includegraphics[width=.19\textwidth]{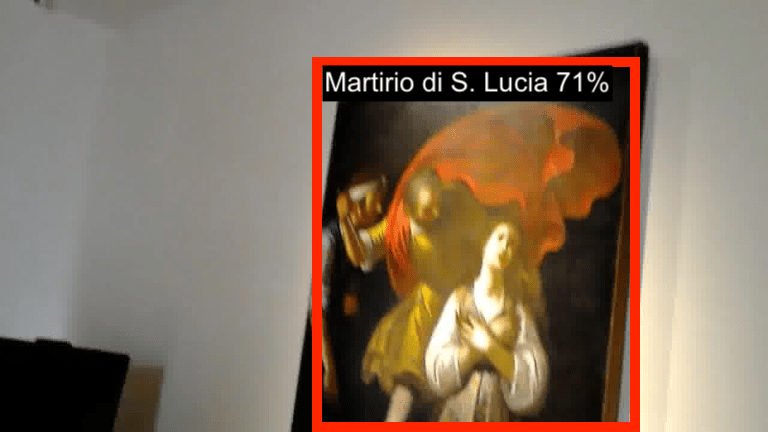}
            \includegraphics[width=.19\textwidth]{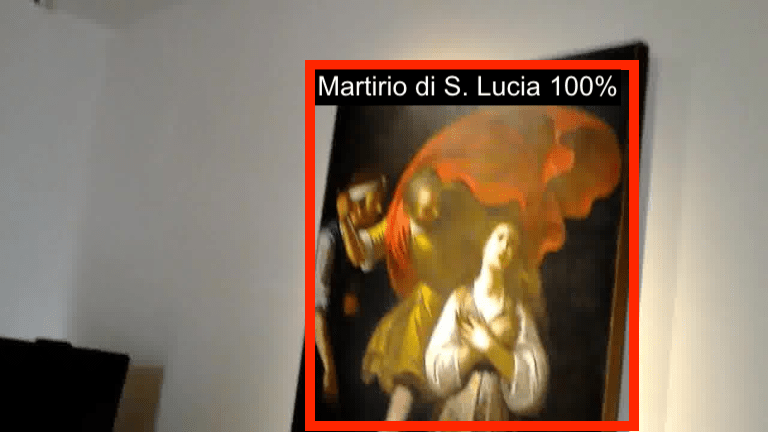}
            
            \vspace{1mm}
            \includegraphics[width=.19\textwidth]{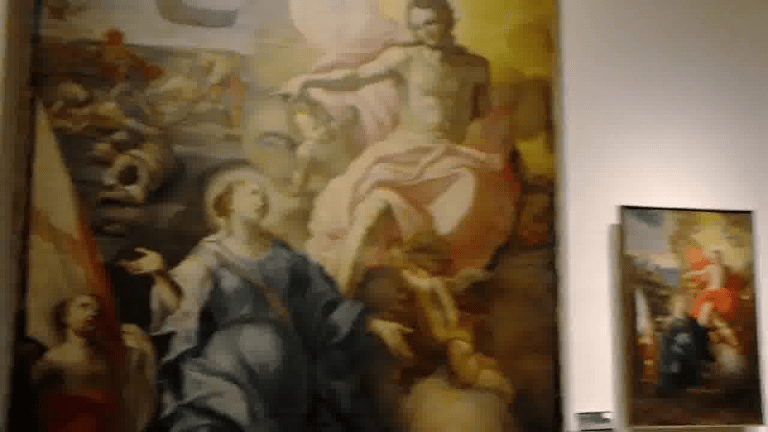}
            \includegraphics[width=.19\textwidth]{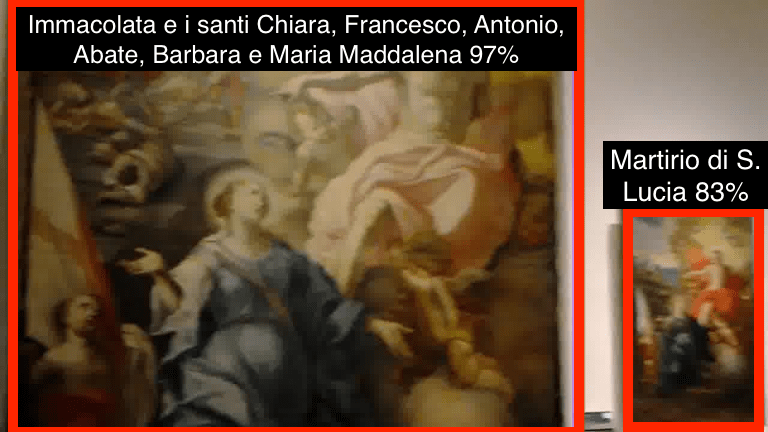}
            \includegraphics[width=.19\textwidth]{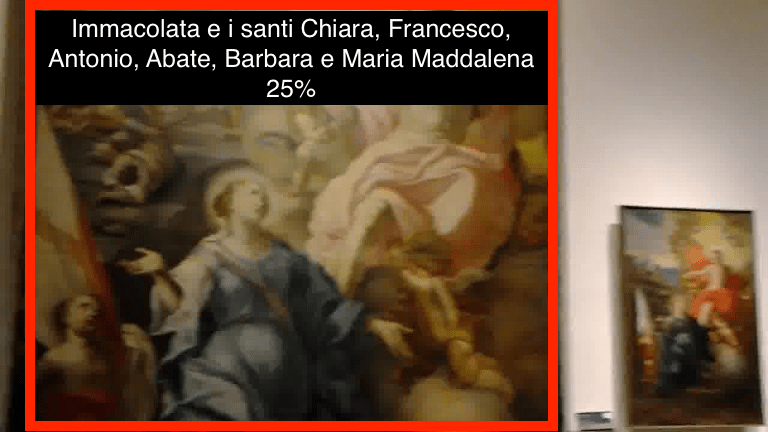}
            \includegraphics[width=.19\textwidth]{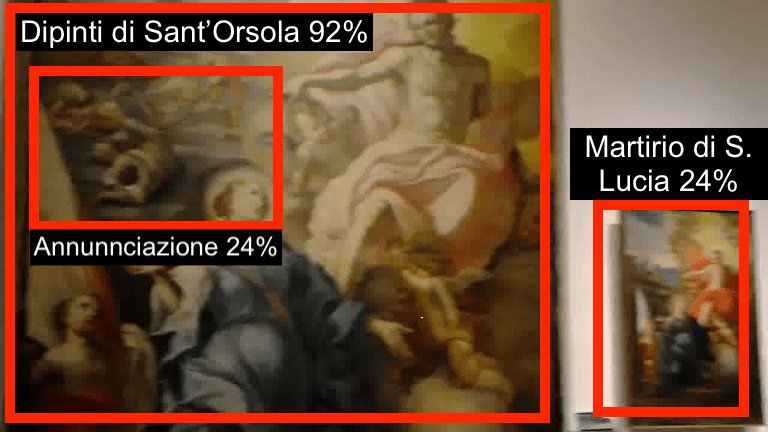}
            \includegraphics[width=.19\textwidth]{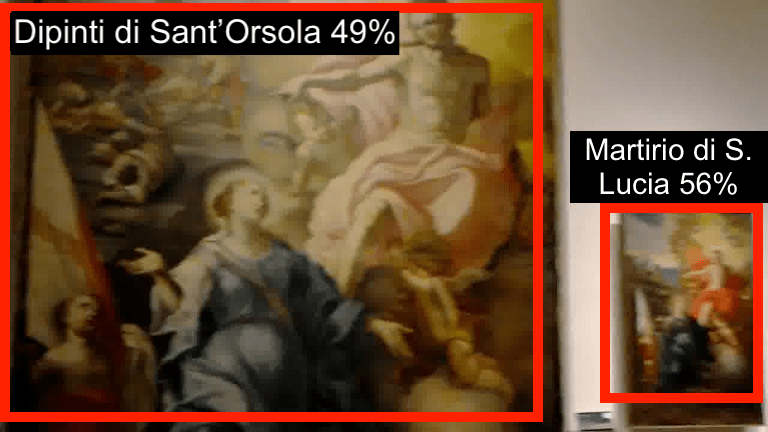}
            
            \vspace{1mm}
            \includegraphics[width=.19\textwidth]{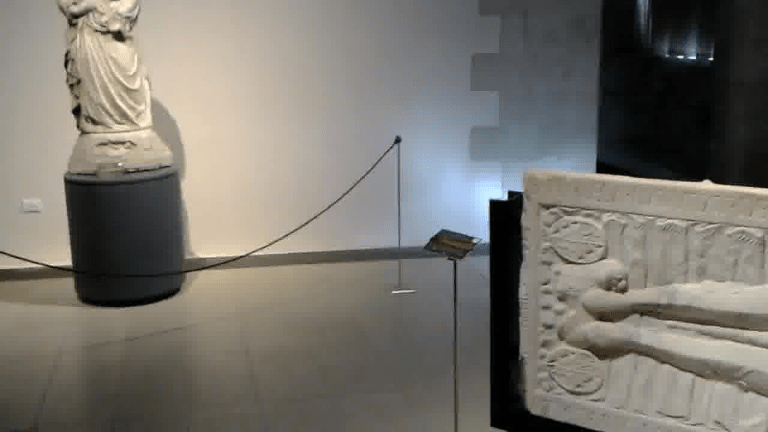}
            \includegraphics[width=.19\textwidth]{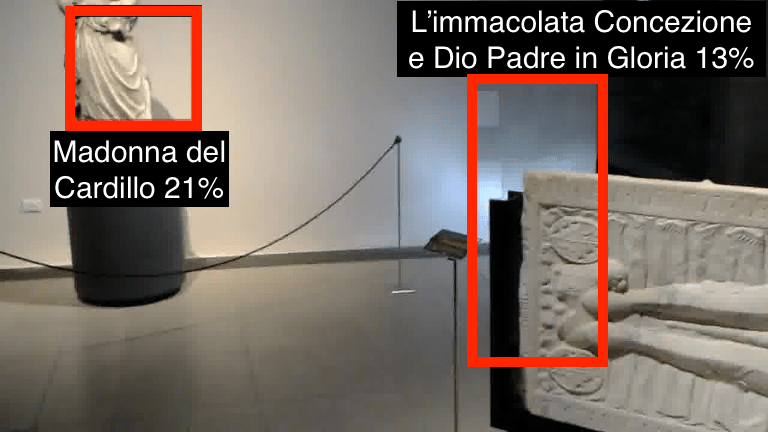}
           \includegraphics[width=.19\textwidth]{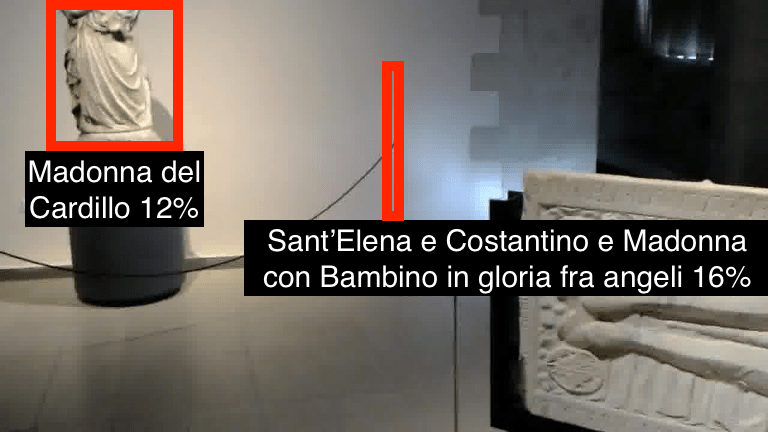}
            \includegraphics[width=.19\textwidth]{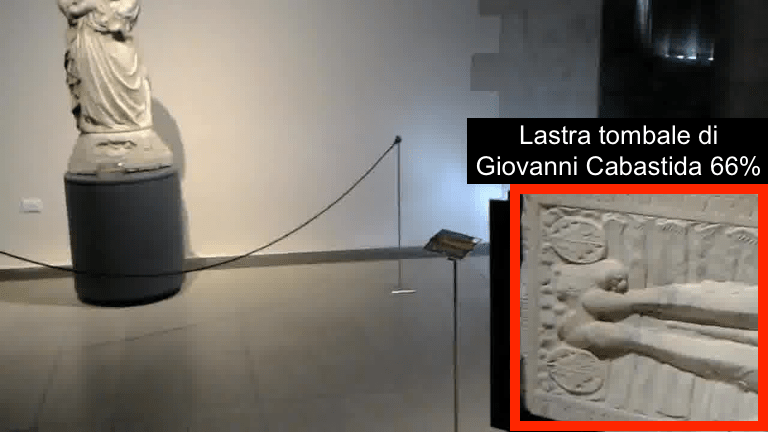}
            \includegraphics[width=.19\textwidth]{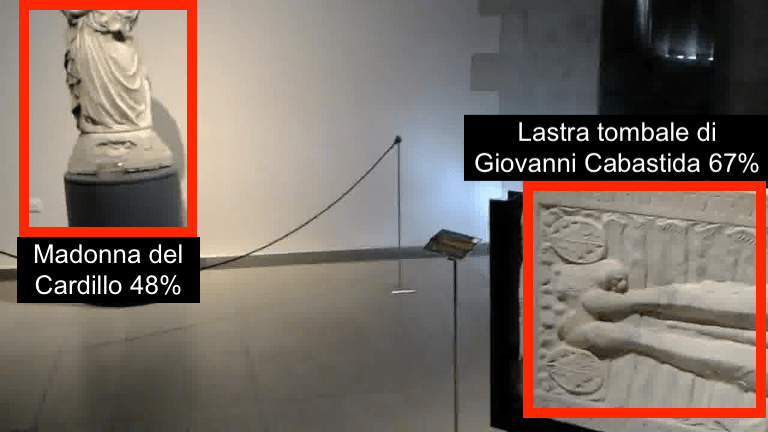}
            
            \vspace{1mm}
            \includegraphics[width=.19\textwidth]{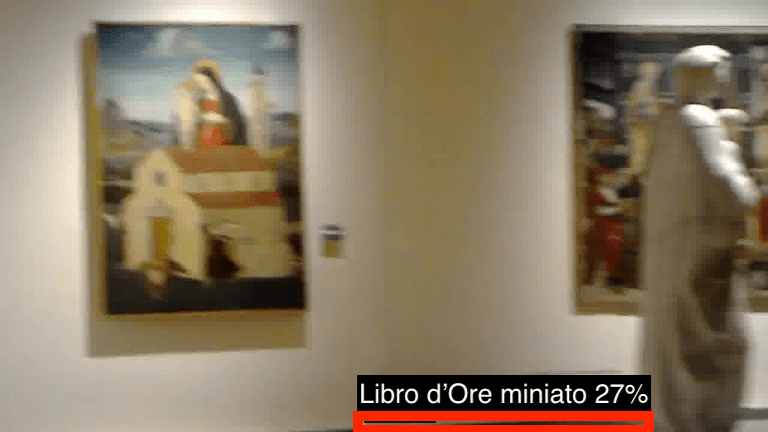}
            \includegraphics[width=.19\textwidth]{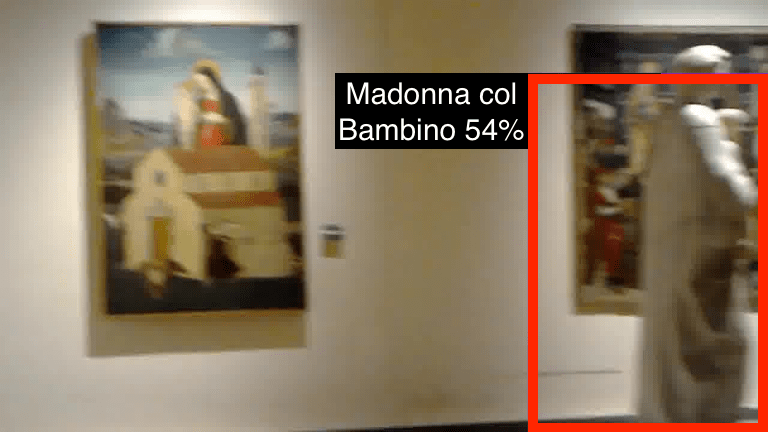}
            \includegraphics[width=.19\textwidth]{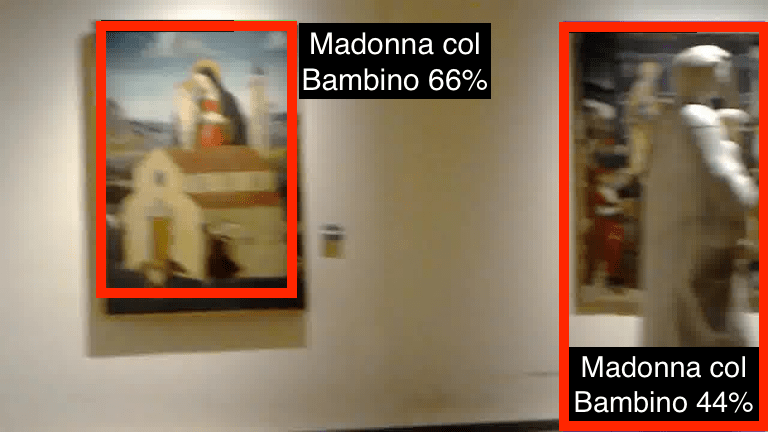}
            \includegraphics[width=.19\textwidth]{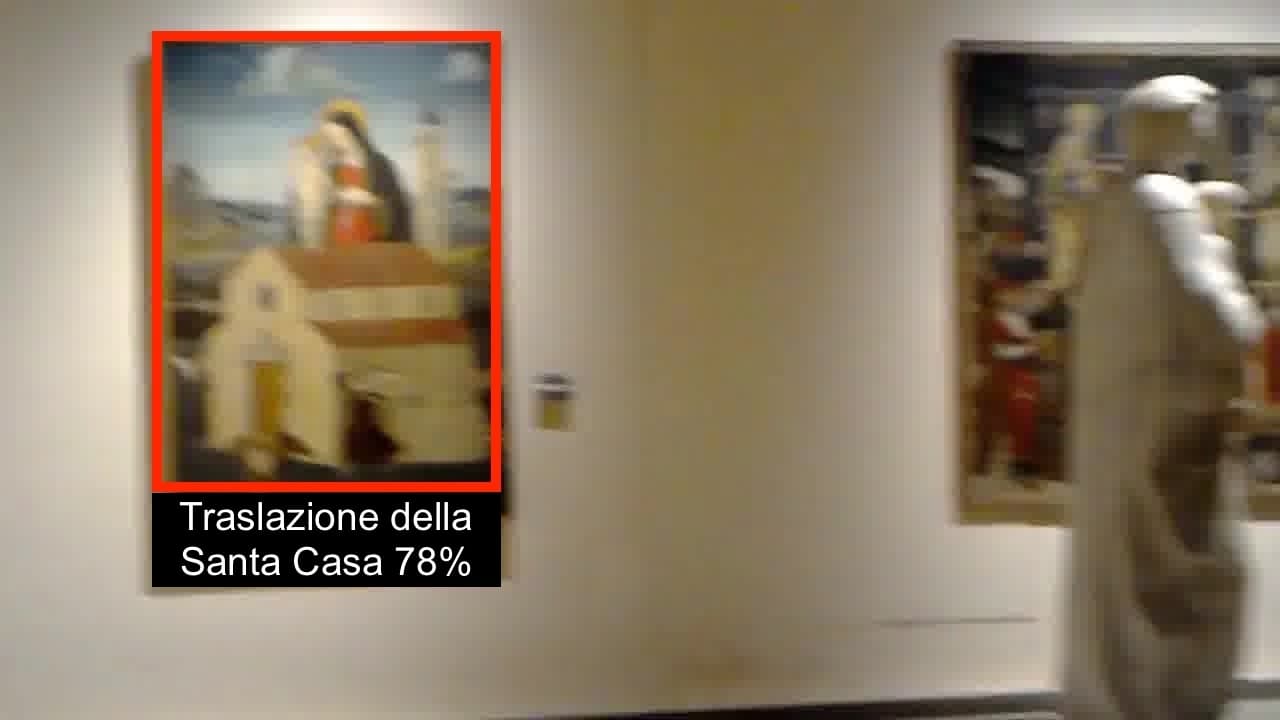}
            \includegraphics[width=.19\textwidth]{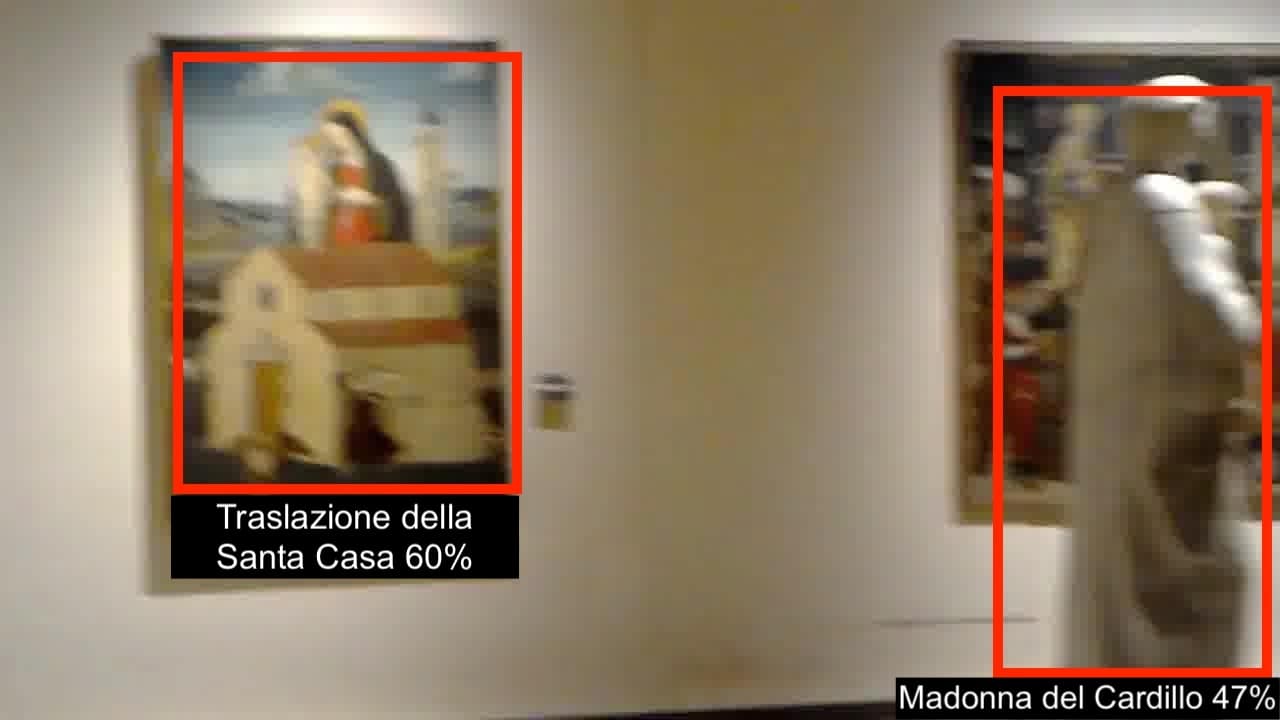}
            
            \vspace{1mm}
            \includegraphics[width=.19\textwidth]{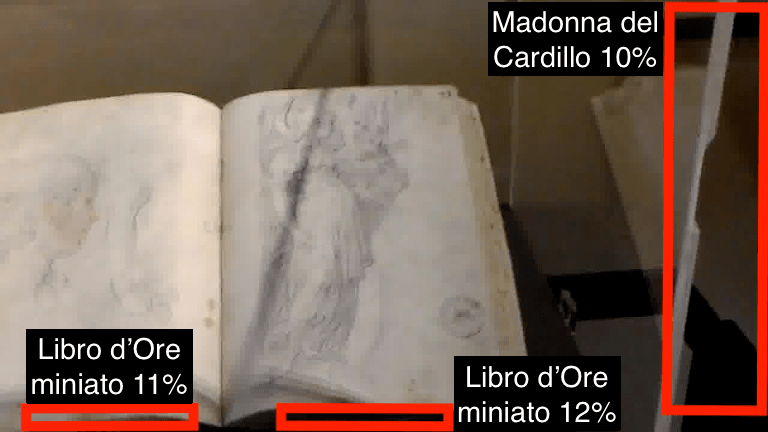}
            \includegraphics[width=.19\textwidth]{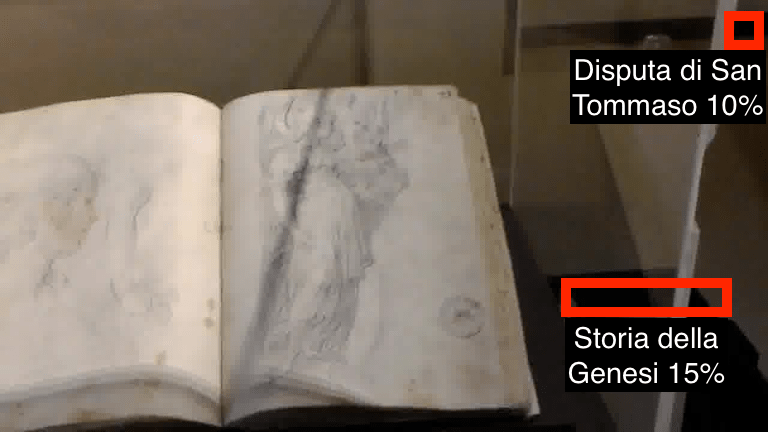}
            \includegraphics[width=.19\textwidth]{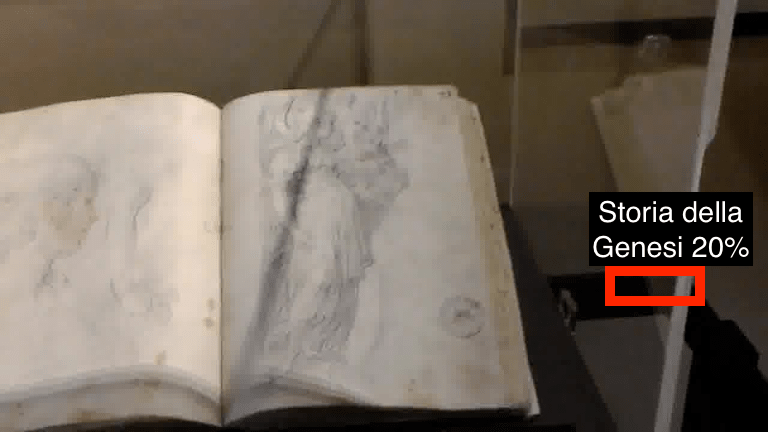}
            \includegraphics[width=.19\textwidth]{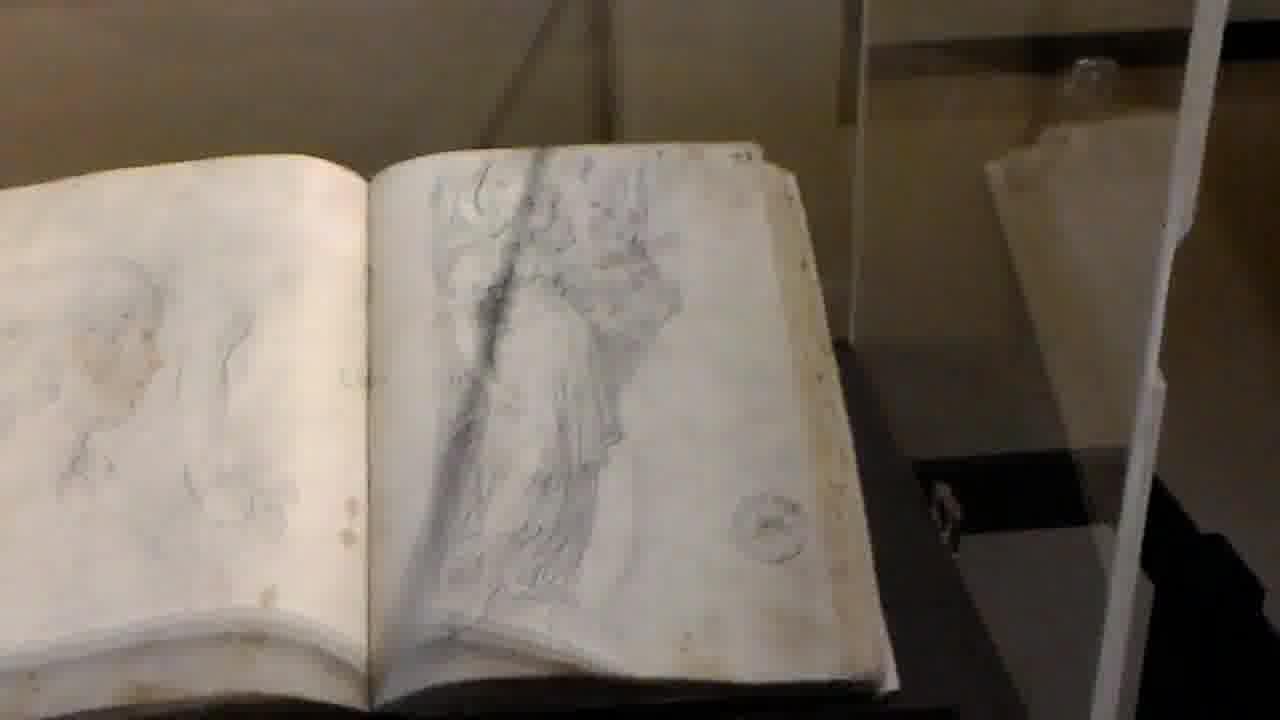}
            \includegraphics[width=.19\textwidth]{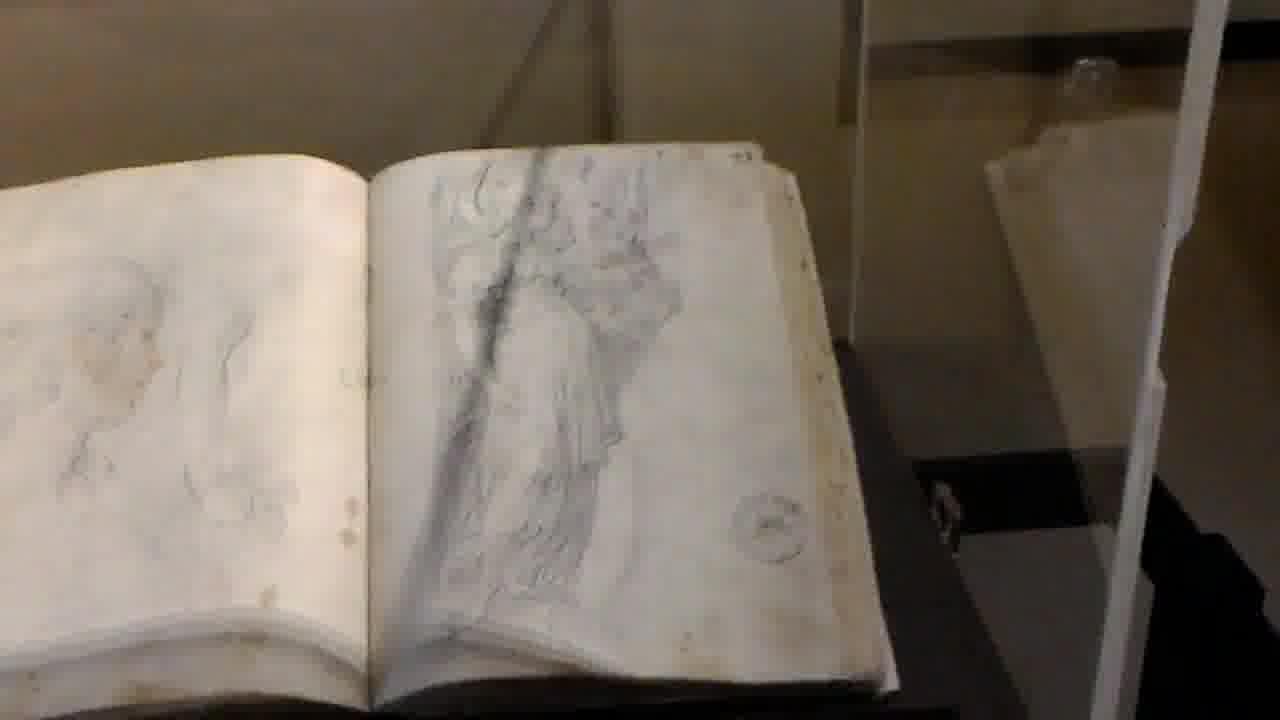}
            
            \vspace{1mm}
            \includegraphics[width=.19\textwidth]{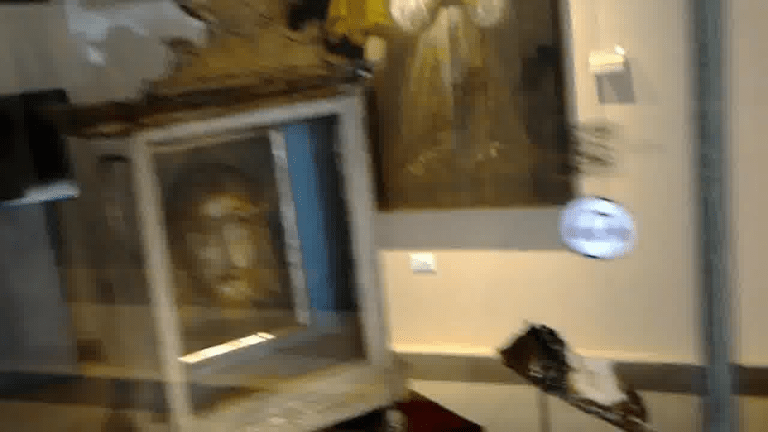}
            \includegraphics[width=.19\textwidth]{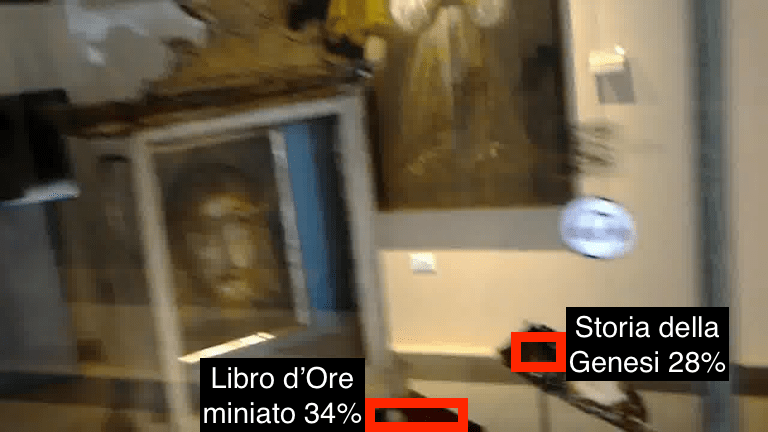}
            \includegraphics[width=.19\textwidth]{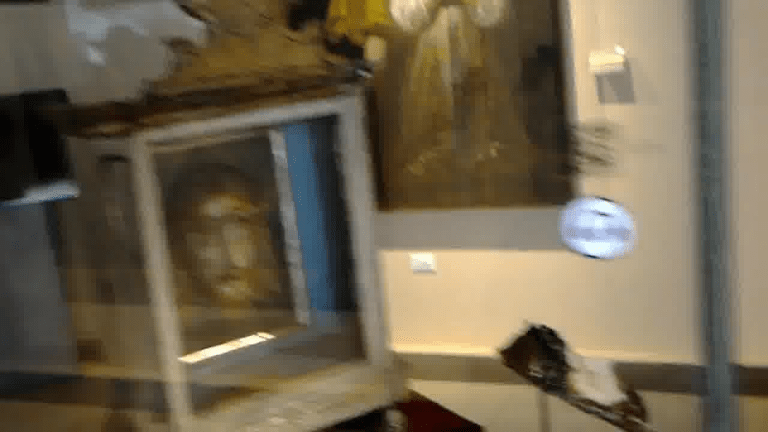}
            \includegraphics[width=.19\textwidth]{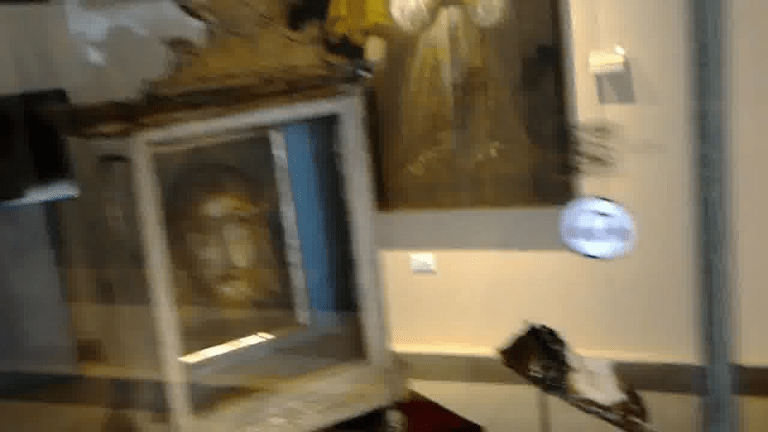}
            \includegraphics[width=.19\textwidth]{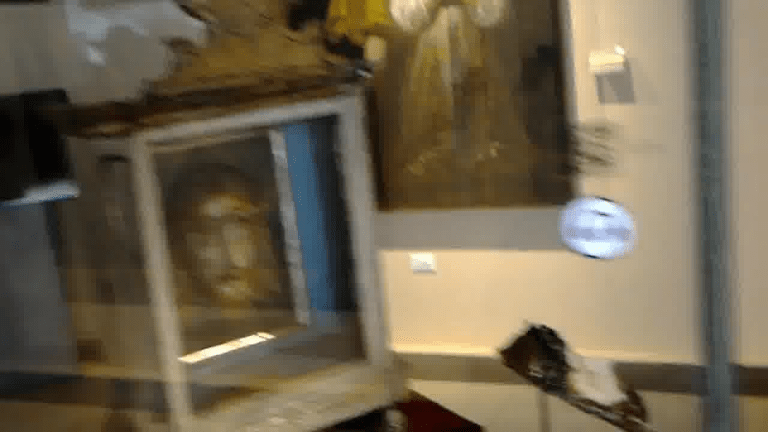}

            \caption{Qualitative results of baseline and feature alignment approaches.}
            \label{fig:qualitativeresult}
\end{figure*}
\begin{figure*}[t!]
            \centering
            \begin{minipage}{.19\textwidth}
            \centering
             Faster RCNN + CycleGAN\\
            \end{minipage}
            \begin{minipage}{.19\textwidth}
            \centering
             DA-Faster RCNN + CycleGAN\\
            \end{minipage}
            \begin{minipage}{.19\textwidth}
            \centering
             Strong-Weak + CycleGAN\\
            \end{minipage}
            \begin{minipage}{.19\textwidth}
            \centering
             RetinaNet + CycleGAN\\
            \end{minipage}
            \begin{minipage}{.19\textwidth}
            \centering
             DA-RetinaNet + CycleGAN\\
            \end{minipage}
            
            \vspace{1mm}
            \includegraphics[width=.19\textwidth]{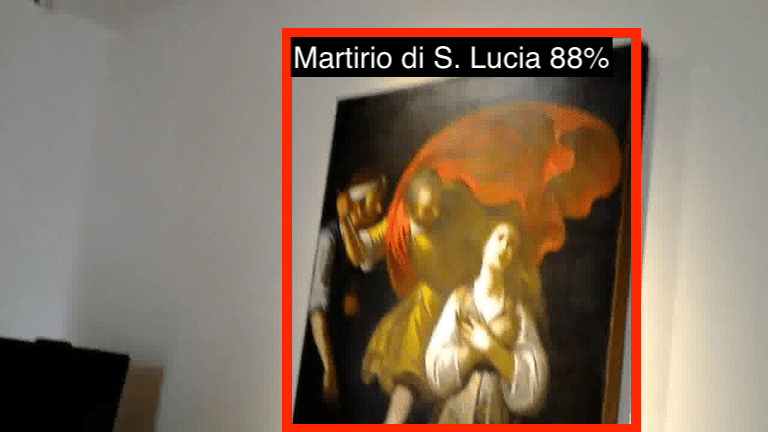}
            \includegraphics[width=.19\textwidth]{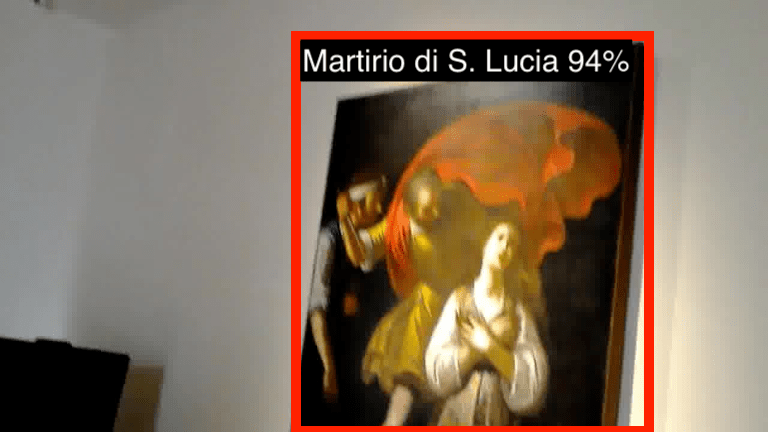}
            \includegraphics[width=.19\textwidth]{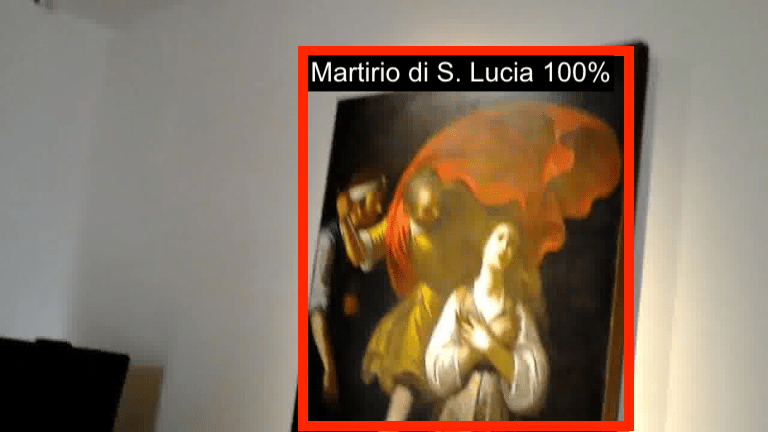}
            \includegraphics[width=.19\textwidth]{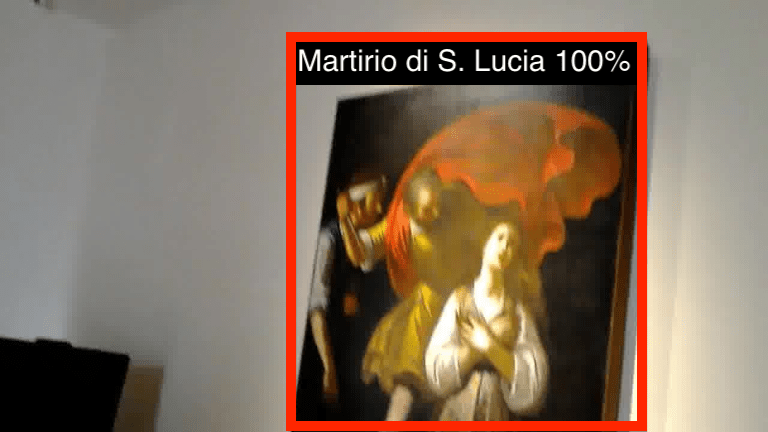}
            \includegraphics[width=.19\textwidth]{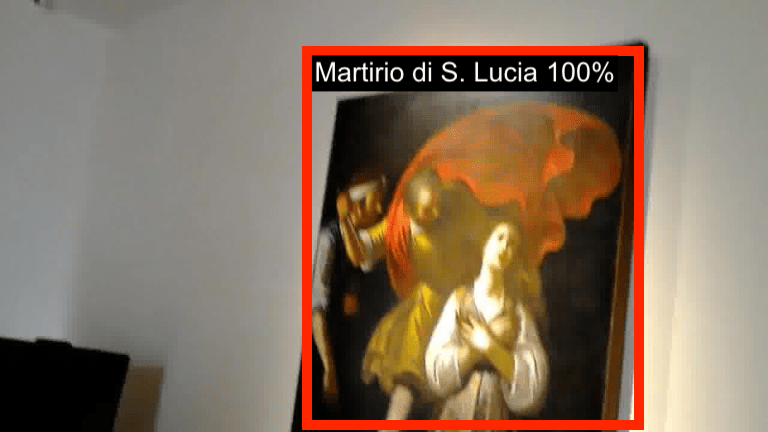}
            
            \vspace{1mm}
            \includegraphics[width=.19\textwidth]{images/result2/fasternoadapt2.png}
            \includegraphics[width=.19\textwidth]{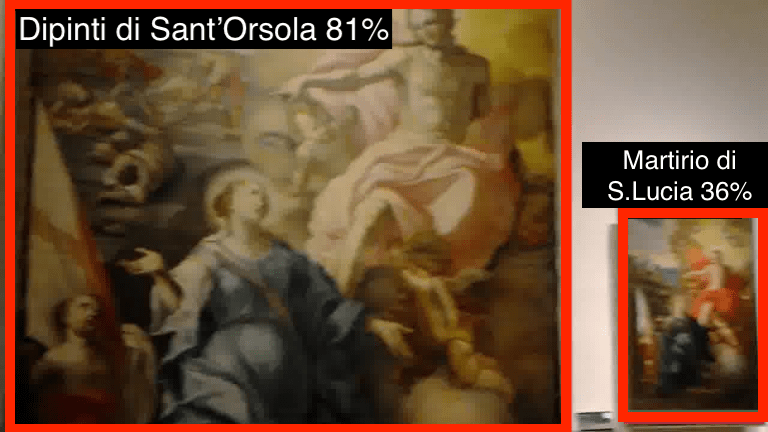}
            \includegraphics[width=.19\textwidth]{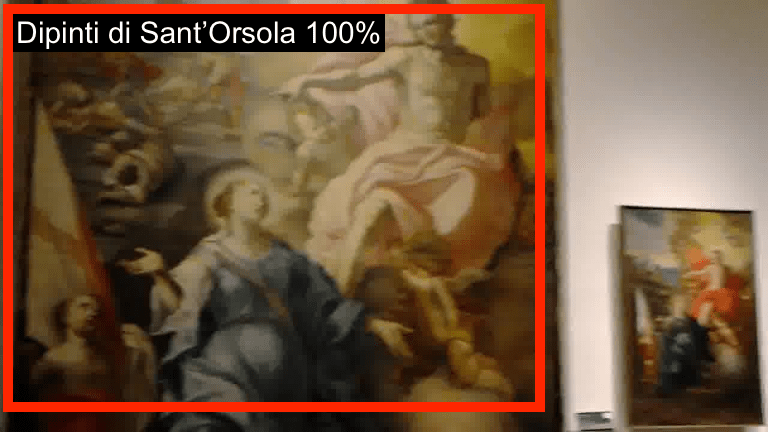}
            \includegraphics[width=.19\textwidth]{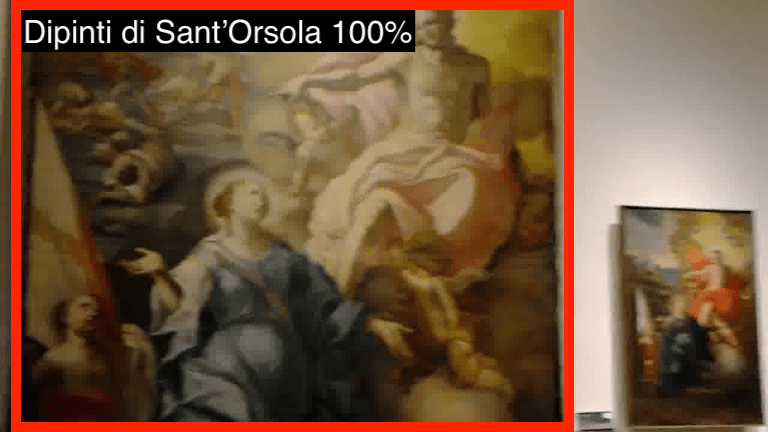}
            \includegraphics[width=.19\textwidth]{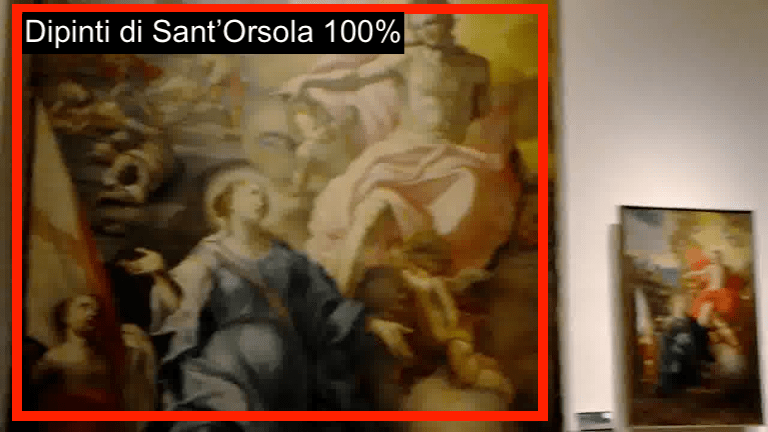}
            
            \vspace{1mm}
            \includegraphics[width=.19\textwidth]{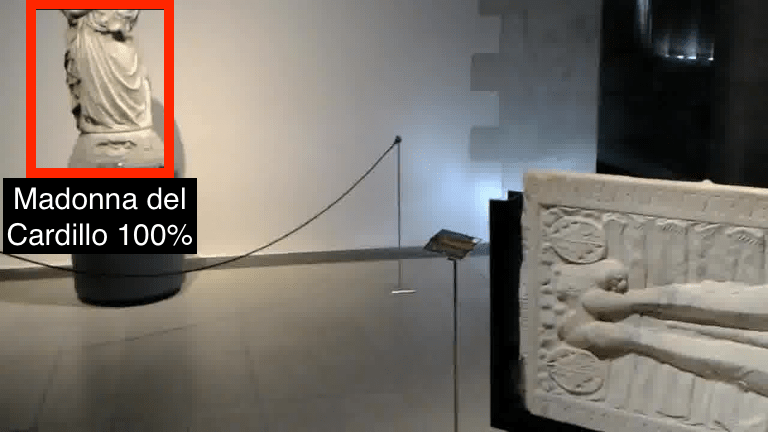}
            \includegraphics[width=.19\textwidth]{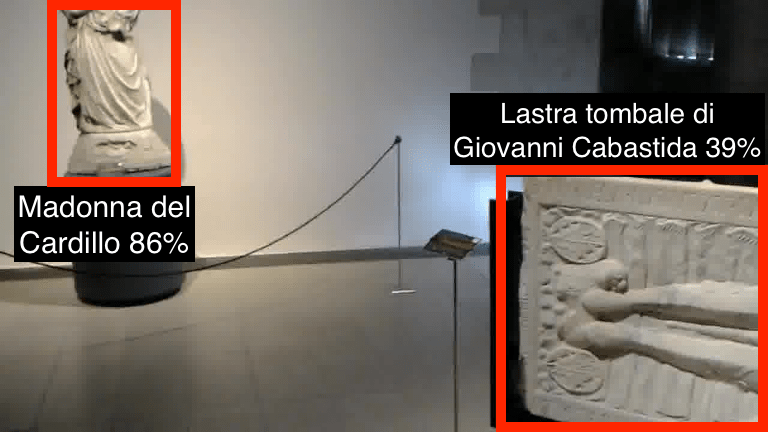}
           \includegraphics[width=.19\textwidth]{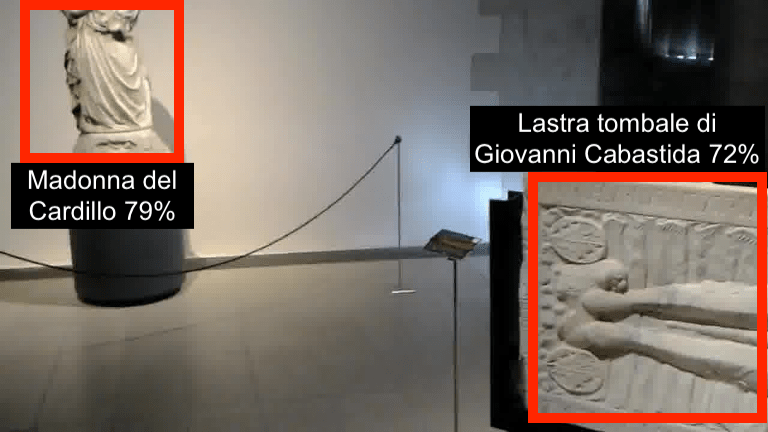}
            \includegraphics[width=.19\textwidth]{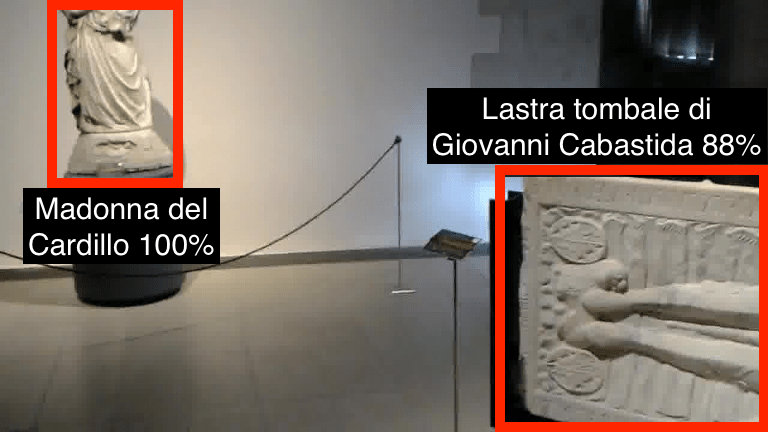}
            \includegraphics[width=.19\textwidth]{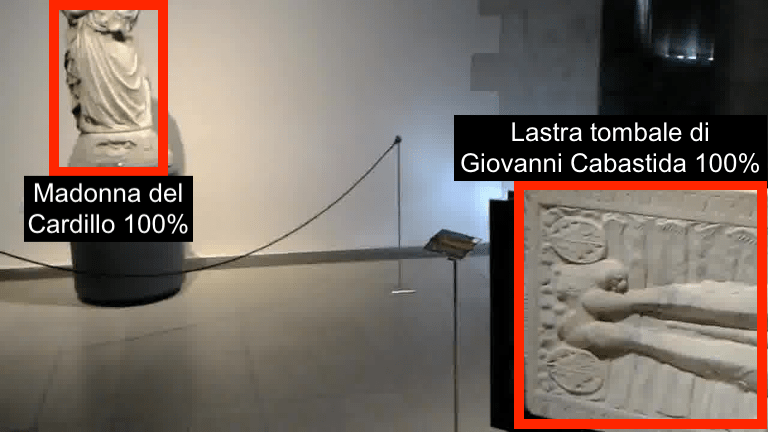}
            
            \vspace{1mm}
            \includegraphics[width=.19\textwidth]{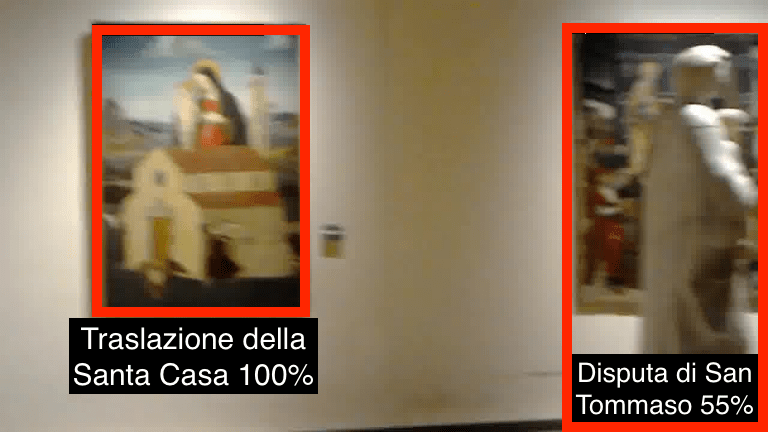}
            \includegraphics[width=.19\textwidth]{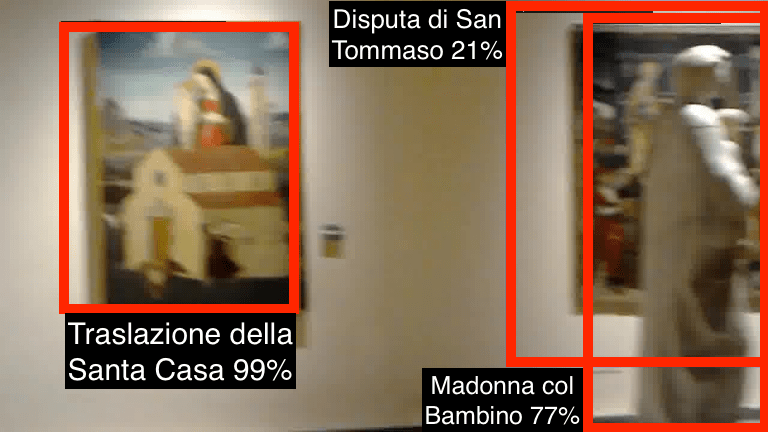}
            \includegraphics[width=.19\textwidth]{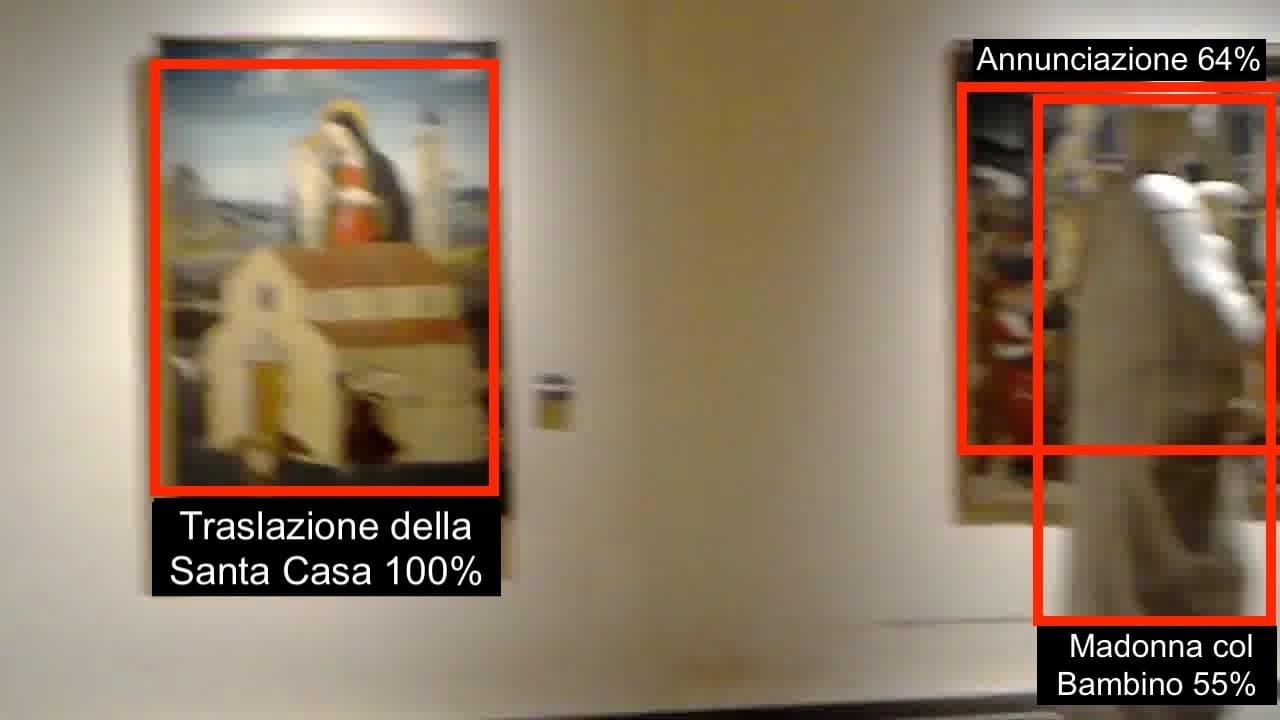}
            \includegraphics[width=.19\textwidth]{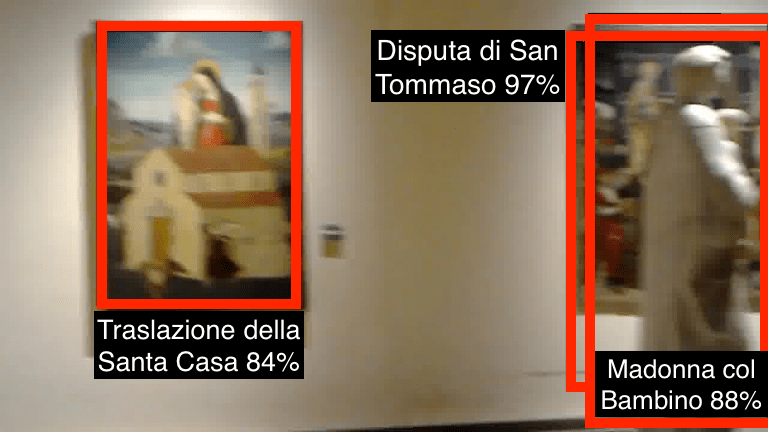}
            \includegraphics[width=.19\textwidth]{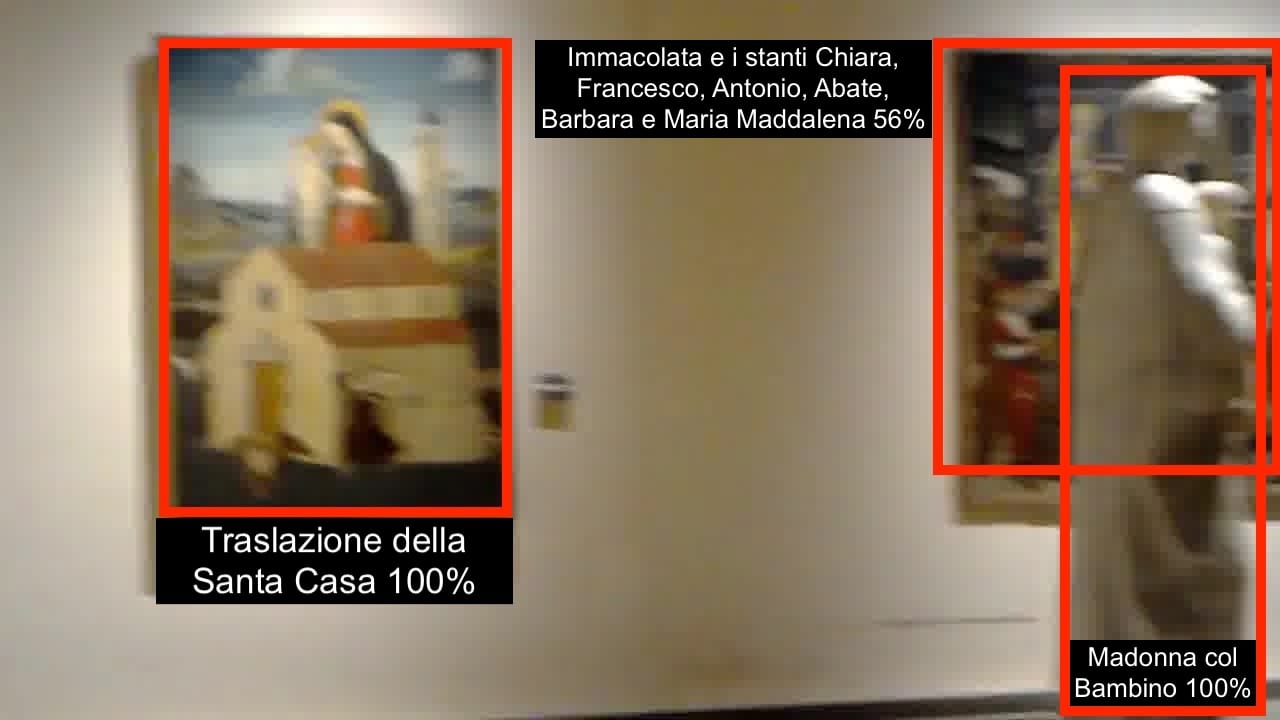}
            
            \vspace{1mm}
            \includegraphics[width=.19\textwidth]{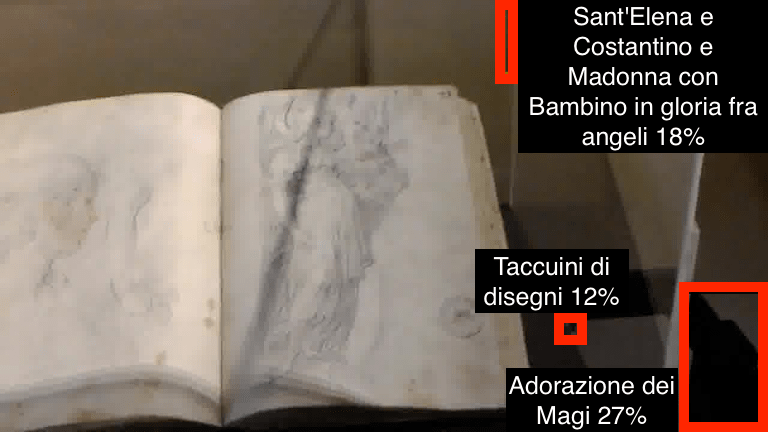}
            \includegraphics[width=.19\textwidth]{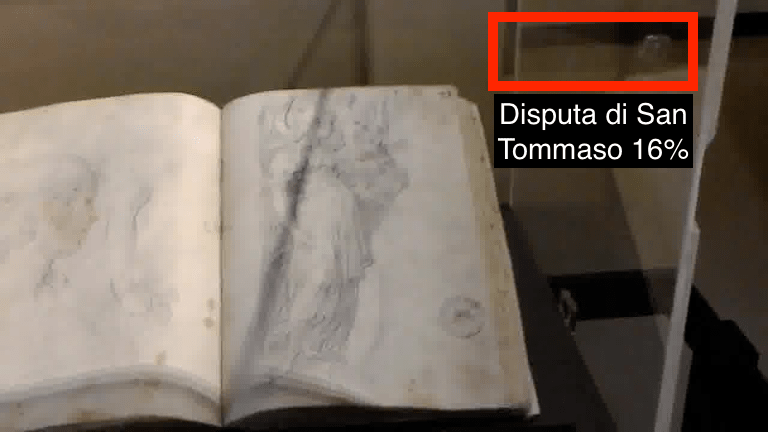}
            \includegraphics[width=.19\textwidth]{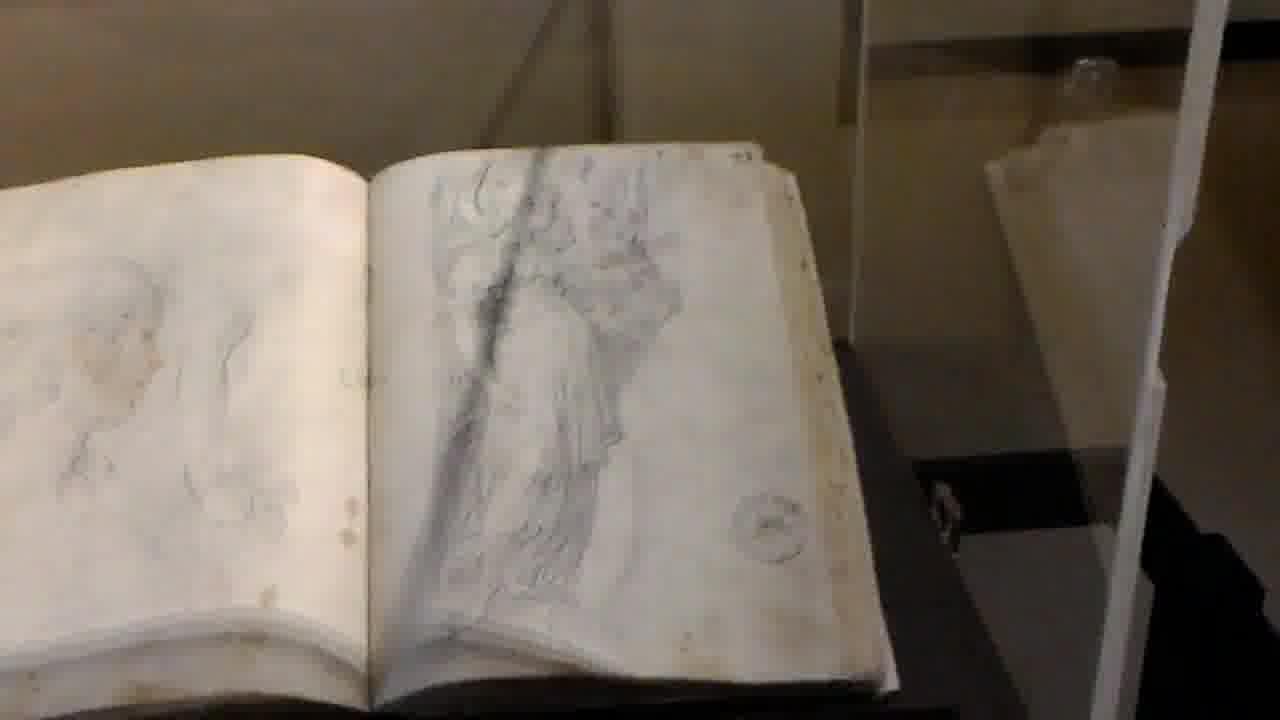}
            \includegraphics[width=.19\textwidth]{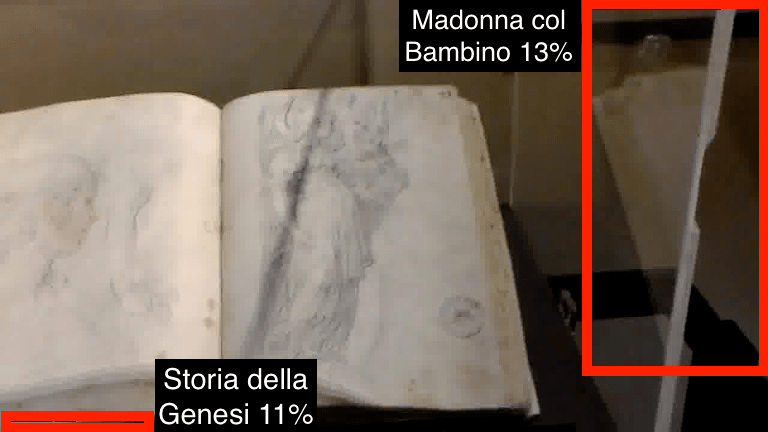}
            \includegraphics[width=.19\textwidth]{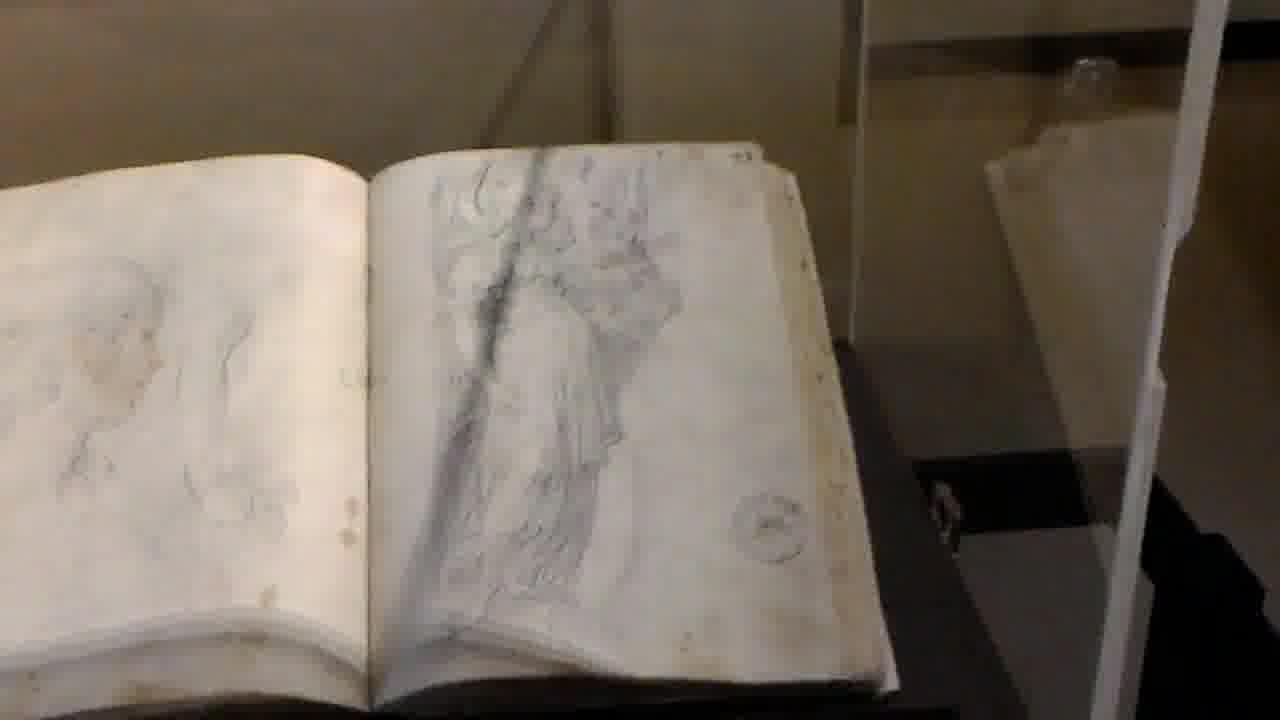}
            
            \vspace{1mm}
            \includegraphics[width=.19\textwidth]{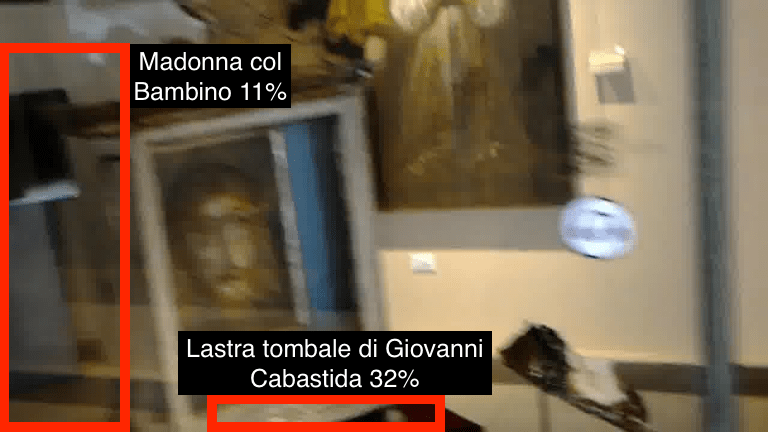}
            \includegraphics[width=.19\textwidth]{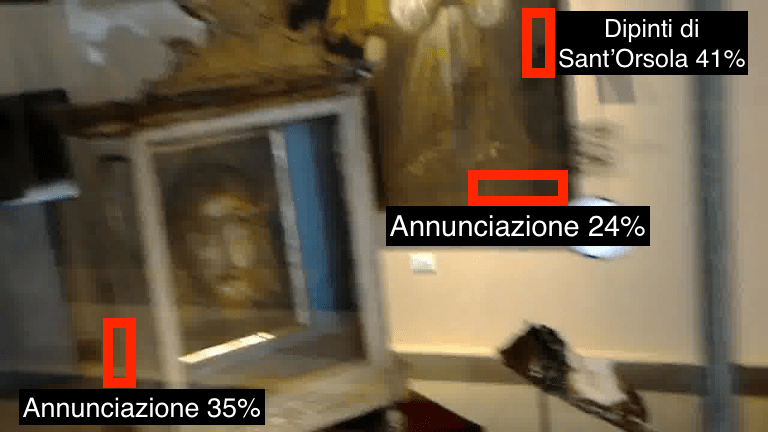}
            \includegraphics[width=.19\textwidth]{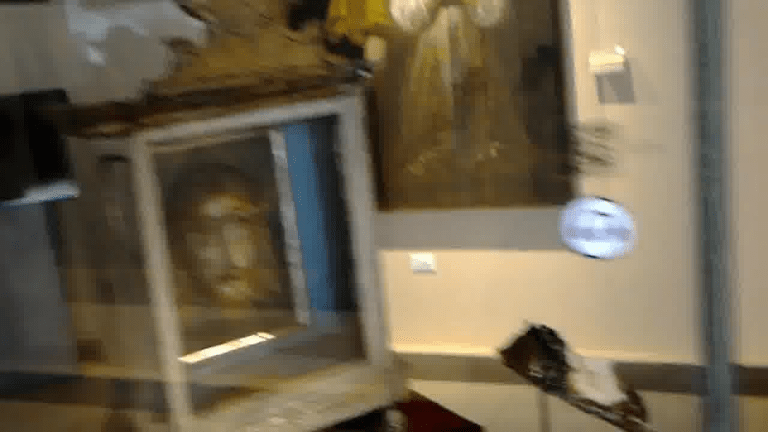}
            \includegraphics[width=.19\textwidth]{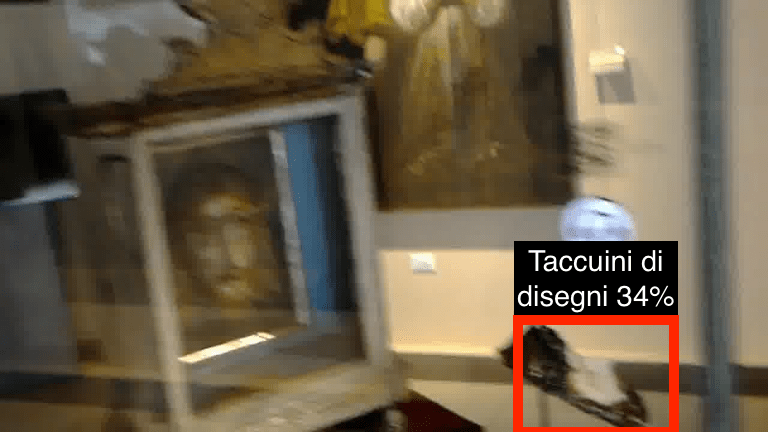}
            \includegraphics[width=.19\textwidth]{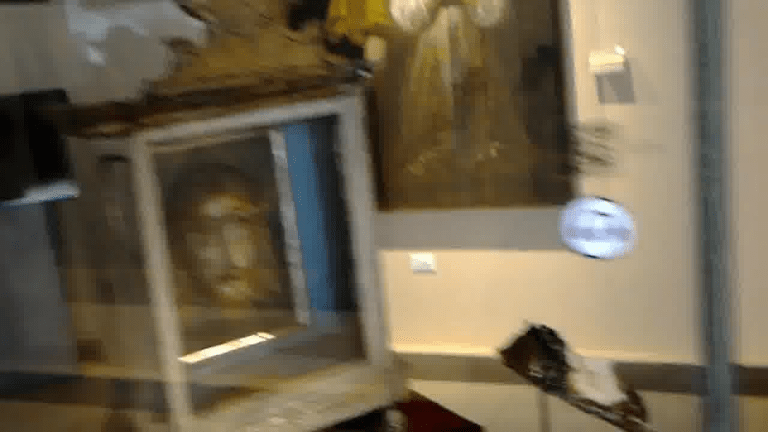}

            \caption{Qualitative results of the baseline and feature alignment combined with CycleGAN.}
            \label{fig:qualitativeresult2}
\end{figure*}

Table~\ref{tab:summary} summarizes all the performances of the analyzed methods with respect to the considered adaptation techniques. The table confirms that the proposed DA-RetinaNet achieves better performance than the compared methods. In particular, considering only feature alignment techniques, our architecture increases performance by about 5\% when compared to Faster RCNN with CycleGAN, and by 6\% compared to Strong-Weak. Again, our method increases the performance of a standard RetinaNet with CycleGAN by about 2.5\% (55.54\% vs 58.01\%). Figure~\ref{fig:qualitativeresult} shows qualitative result of the baseline and the models based on feature alignment. Faster RCNN does not detect any object and in some cases its predictions are false positive. DA-Faster RCNN and RetinaNet correctly detect objects in the ``easy" examples (first two rows), with some misclassification problems when there are more objects and occlusions (last three rows). Strong-Weak and DA-RetinaNet are more accurate in detection but they still produce some false positive and false negative predictions. Figure~\ref{fig:qualitativeresult2} reports the qualitative results of the previous five methods combined with CycleGAN to translate images from synthetic to real. Faster RCNN and DA-Faster RCNN have similar results to DA-RetinaNet but they have much more false positive detections. Strong-Weak and RetinaNet combined with CycleGAN correctly detect the objects of the first three rows. Strong-Weak is less accurate than RetinaNet but has less false positive detections. DA-RetinaNet combined with CycleGAN perfectly detects artworks in the first four rows with only a misclassification in the fourth rows behind the statue. As can be noted, even these models are not able to detect object in the last two rows. Possible reasons are: 1) bad translation results from synthetic to real, 2) few synthetic object are not similar to their real counterpart, 3) some synthetic objects are similar to each other (e.g. some books). 
\subsection{Analysis of Computational Resources}
\label{computational_re}
\begin{table}[t!]
\caption{Training times required by the models.}
\label{time}
\centering
\begin{tabular}{|c||c|}
\hline
Model & Hours (Days) \\
\hline
RetinaNet (12K iterations) & $\sim 10\; (\sim 0.5)$\\
\hline
RetinaNet (62K iterations) & $\sim 65 \; (\sim 3)$\\
\hline
DA-RetinaNet & $\sim 67\; (\sim 3)$\\
\hline
Faster RCNN (62K iterations) &  $\sim 131 \; (\sim 5.5)$\\
\hline
DA-Faster RCNN & $\sim 142 \; (\sim 6)$\\
\hline
Strong-Weak & $\sim 147\; (\sim 6)$\\
\hline
CycleGAN &  $\sim 1470 \; (\sim 61)$\\
\hline
CycleGAN + RetinaNet &  $\sim 1535 \; (\sim 64)$\\
\hline
CycleGAN + DA-RetinaNet &  $\sim 1537 \; (\sim 64)$\\
\hline
CycleGAN + Faster RCNN &  $\sim 1601 \; (\sim 66)$\\
\hline
CycleGAN + DA-Faster RCNN &  $\sim 1612 \; (\sim 67)$\\
\hline
CycleGAN + Strong-Weak &  $\sim 1617 \; (\sim 67)$\\
\hline
\end{tabular}
\end{table}
Table~\ref{time} shows the training times required by the algorithms using a single NVIDIA Tesla K80. We use the same batch size for each object detector to evaluate the training times. Training CycleGAN for 60 epochs required 61 days in the considered settings. Methods based on feature alignment require from 3 to 6 days depending on the considered object detector. In particular, DA-Faster RCNN, Strong-Weak and DA-RetinaNet have only a small computational overhead given by the presence of the discriminators. However, even if these methods required less time when compared to CycleGAN, they have limited performance when compared to their counterparts who make use of image-to-image translation (e.g. DA-RetinaNet: 31.04 \% vs 58.01 \%, Strong-Weak: 25.12 \% vs 47.70 \%, DA-Faster RCNN: 12.94 \% vs 33.20 \%). We argue that more attention should be devoted to such approaches in order to minimize training times.
\subsection{Results on Cityscapes Dataset}
To better asses the performance of the proposed method and to understand generalization capability over datasets, we have performed experiments on the Cityscapes dataset~\cite{cordts2016cityscapes}~\cite{sakaridis2018semantic}. To this aim, we trained RetinaNet and DA-RetinaNet for 50K iteration with a learning rate of 0.0002, batch size of 4 and starting from weights pre-trained on ImageNet. Following~\cite{Saito_2019}, we used Cityscapes~\cite{cordts2016cityscapes} as source domain and Foggy-Cityscapes~\cite{sakaridis2018semantic} as target domain. Both dataset have 2975 images in the training set. We reported results on the 500 images of the validation set. Table~\ref{tab:cityscape} reports the results obtained by standard object detector architectures and domain adaptation methods based on feature alignment. The table highlights that standard RetinaNet achieves better performance than Strong-Weak and Diversify and Match by about 6\%. The proposed DA-RetinaNet increases performance by 4\%, 10\%, and 24\% if compared respectively with standard RetinaNet, Strong Weak and Diversify and Match, and DA-Faster RCNN. However there is still a gap between the best results obtained by the proposed architecture and the result of the Oracle which is obtained training and testing RetinaNet on the Foggy Cityscapes dataset, which suggests that there is still room for improvement.
\begin{table}[t!]
\caption{Results adaptation between Cityscapes and Foggy Cityscapes dataset. The performance scores of the methods marked with the ``*” symbol are reported from the authors of their respective papers.}
\label{tab:cityscape}
\centering
\begin{tabular}{|c|c|}
\hline
Model & mAP\\
\hline
Faster RCNN*~\cite{Saito_2019}  & 20.30\%\\
\hline
DA-Faster RCNN*~\cite{chen2018domain}  & 27.60\%\\
\hline
Strong-Weak*~\cite{Saito_2019}  & 34.30\%\\
\hline
Diversify and Match*~\cite{kim2019diversify} & 34.60\%\\
\hline
RetinaNet  & 40.25\%\\
\hline
DA-RetinaNet  & \textbf{44.87\%}\\
\hline
Oracle & 53.46\%\\
\hline
\end{tabular}
\end{table}

\section{Conclusion}
We considered the problem of Unsupervised Domain Adaptation for object detection in cultural. To conduct our study, we created a new dataset consisting of 75244 synthetic images and 2190 real images of 16 artworks, which we publicly release. To better assess generalization of the compared approaches, we have also performed experiment with a dataset related to urban environment.
Experiments showed that the proposed DA-RetinaNet method achieves better performance compared to DA-Faster RCNN and Strong-Weak. 
At the same time, the results obtained by these methods based on feature alignment achieved very poor performance if compared to their counterparts combined with image-to-image translation techniques. 
DA-RetinaNet performed better than others also when combined with CycleGAN. However, using CycleGAN with this dataset required a high computational training cost.
We hope that the proposed dataset will encourage research on this challenging topic and that the proposed DA-RetinaNet will serve as a strong baseline for future works.
\label{conclusion}
\section*{Acknowledgments}
This research is supported by the project VALUE - Visual Analysis for Localization and Understanding of Environments (N. 08CT6209090207 - CUP G69J18001060007) - PO FESR 2014/2020 - Azione 1.1.5., by Piano di incentivi per la ricerca di Ateneo 2020/2022 (Pia.ce.ri.) Linea 2 - University of Catania, and by MIUR AIM - Attrazione e Mobilit\`a Internazionale Linea 1 - AIM1893589 - CUP E64118002540007.
\bibliography{main}

\end{document}